\documentclass[smallcondensed]{svjour3}

\usepackage{graphicx}
\usepackage{multirow}
\usepackage{hhline}
\usepackage{enumitem}
\usepackage{amssymb,amsmath,amsfonts,enumerate,float
}
\usepackage{graphicx}
\usepackage{subcaption}
\usepackage[ruled,norelsize,linesnumbered]{algorithm2e}
\usepackage{bbm}
\usepackage{natbib}
\begin{document}

\title{Variational Auto-encoder Based Bayesian Poisson Tensor Factorization for Sparse and Imbalanced Count Data
}


\author{Yuan Jin \and Ming Liu \and Yunfeng Li \and Ruohua Xu \and Lan Du \and Longxiang Gao \and Yong Xiang
}


\institute{Yuan Jin \at
              Faculty
of Information Technology, Monash University, Melbourne, Victoria 3800, Australia \\
              \email{yuan.jin@monash.edu}           
           \and
           Lan Du \at Faculty
of Information Technology, Monash University, Melbourne, Victoria 3800, Australia \\ \and Ming Liu \and Longxiang Gao \and Yong Xiang \at School
of Information Technology, Deakin University, Melbourne,
Victoria 3125, Australia
\and Yunfeng Li \and Ruohua Xu \at Beijing Shandesitong Technology, Haidian Qu, Beijing
100081, China
}

\date{Received: date / Accepted: date}

\maketitle

\begin{abstract}
Non-negative tensor factorization models enable predictive analysis on count data. Among them, Bayesian Poisson-Gamma models can derive full posterior distributions of latent factors and are less sensitive to sparse count data. However, current inference methods for these Bayesian models adopt restricted update rules for the posterior parameters. They also fail to share the update information to better cope with the data sparsity. Moreover, these models are not endowed with a component that handles the imbalance in count data values. In this paper, we propose a novel variational auto-encoder framework called VAE-BPTF which addresses the above issues. It uses multi-layer perceptron networks to encode and share complex update information. The encoded information is then reweighted per data instance to penalize common data values before aggregated to compute the posterior parameters for the latent factors. Under synthetic data evaluation, VAE-BPTF tended to recover the right number of latent factors and posterior parameter values. It also outperformed current models in both reconstruction errors and latent factor (semantic) coherence across five real-world datasets. Furthermore, the latent factors inferred by VAE-BPTF are perceived to be meaningful and coherent under a qualitative analysis.
\end{abstract}

\section{Introduction}\label{sec:introduction}
In this paper, we focus on improving the performance of \textit{Bayesian Poisson tensor factorization} (BPTF). In terms of BPTF, it imposes \textit{Gamma} distributions as priors over its latent factors. These factors then form the instance-wise rates for a Poisson likelihood over data observations. BPTF adopts two types of inference frameworks to compute the posterior shape and rate for its Gamma latent factors: \textit{Gibbs sampling} and \textit{variational inference}. Both of them rely on the \textit{auxiliary variable augmentation} technique to facilitate their computation. This technique is based on the Poisson-Gamma \textit{conjugacy}. It exploits the fact that a sum of auxiliary Poisson variables with respective rates is itself a Poisson with the rate equal to the sum of the auxiliaries' rates. 

Despite its importance, the augmentation technique, however, increases the computation overhead due to the additional sampling procedures/updates on the auxiliary Poisson variables. Moreover, the updates on each latent factor are independent and thus fail to utilize the information from each other. This limits the performance of BPTF when it encounters sparse tensors. In this case, we want the data information for latent factors to be shared to enhance the inference of their posterior distributions.

A common strategy to share the data information is to treat the parameters of the posterior distributions of latent factors as regression models. The regression coefficients are learned to map similar data patterns into values in close proximity in the latent space. Naturally, the mapping is non-linear. This motivates us to use \textit{artificial neural networks}, which can fit complex mapping functions, to estimate the posterior distribution parameters for the latent factors. \textit{Variational auto-encoder} (VAE)~\citep{kingma2013auto} provides the foundation to achieve all of the above. It links variational inference of posterior distribution parameters with multi-layer perceptron (MLP) networks. 

In this paper, we propose a novel factorization framework that combines BPTF with VAE. It conducts \textit{mean-field} variational inference for latent factors under each mode of a tensor using \textit{mode}-specific MLP networks. These networks, acting as the encoders, compute the posterior shapes and rates for Gamma latent factors under each mode. Furthermore, the encoder network for a latent factor from a particular mode takes in both its data and the latent factors from the other modes. This encoding style differs from that of classifical VAE. In classifical VAE, only the data is encoded but not the other latent factors that also contribute to generating the data. Unlike VAE, our framework has instance-wise inputs, each comprising a data instance and latent factors that generate the data instance. This allows the input dimension of our framework to grow linearly with the number of modes rather than the number of possible entries in the tensor as in the VAE. 

Our framework estimates the posterior shape and rate for a Gamma latent factor based on its associated instance-wise inputs. Each input may contribute differently to the estimation. Thus, the estimation is done by \textit{summing softplus activation} of the outputs from the corresponding encoder. The softplus function yields a value close to zero when an input contributes little to the estimation. The sum-of-softplus operation balances the extent to which instance-wise contributions are sparsified against numerical stability for which the shape and rate must be greater than zero. 

Our framework also handles the data value imbalance problem for BPTF. In a typical application with BPTF, word counts are collected from publication databases as a four-way tensor. Its modes correspond to authors, words, years and publication venues. Each entry of this tensor records the number of times a word appears in a scholar's articles published at a venue in a particular year. In this case, the majority of non-zero entries in this tensor will be one (i.e. most words occurred only once). BPTF is likely to be overwhelmed by the influence of such imbalance in its inference.

To solve this issue, our framework further weighs each softplus activation by how far their input data values are from the average (or most frequent) value. A data value farther from the average tends to have more useful information in revealing how its associated latent factors are distributed. Thus, more weights should be given to its corresponding softplus activation to increase its contribution in posterior parameter estimation.

Experimental results show that our framework outperforms several state-of-the-art factorization techniques on predicting missing values for non-negative multi-way tensors. Moreover, we show that our framework has more potential of learning meaningful and coherent latent factor structures for the tensors. 

\section{Related Work}
Real-world data is always generated as the outcomes of some events 
which can be organized into \textit{multi-way tensors}. 
Given the observed event outcomes as observed data entries in a tensor, it is important to conduct predictive analysis for the unobserved outcomes. 
They are presented as missing values in the tensor. The core of the predictive analysis is to uncover the underlying latent structures that have generated the observed data entries. It is then straightforward to predict the missing values, assuming that they share the same latent structures with the observed ones.

\subsection{Non-negative Tensor Factorization}
Tensor factorization (TF) techniques~\citep{kolda2009tensor} provide effective means to uncover the latent structures. They decompose a tensor into latent factor matrices specific to its modes. Latent factors represent the underlying characteristics of each element within the corresponding mode.

A significant application of the TF techniques is the predictive analysis for \textit{non-negative integer data}. This type of data is widespread across many areas such as recommendation, publication, and crowd-sourcing systems, etc. The two most common forms in which this type of data are observed are the rating and count. The former is of particular interest to the area of recommendation systems where data takes the form of integers under small scales (e.g. 1-5). The latter does not impose any scale constraint on data values.

\textit{Non-negative} tensor factorization (NTF)~\citep{welling2001positive,shashua2005non,friedlander2008computing,chi2012tensors} was developed to decompose non-negative integer data. Originally, it was formulated as a constrained minimization problem. The objective function can be constructed based on various measures of the discrepancy between the observed and predicted data values. Typical choices of the measure are Euclidean distance~\citep{shashua2005non,kolda2009tensor} and Kullback-Leibler (KL) divergence~\citep{welling2001positive,chi2012tensors}. Each prediction is based on latent factors constrained to be non-negative. Classical NTF finds locally optimal point estimates for the latent factors. In~\citep{chi2012tensors}, the authors used Poisson likelihood to model sparse count data and maximized its logarithm which turns out to be the KL divergence.

Unlike the previous non-Bayesian point estimation NTF, Bayesian NTF infers the full posterior distributions of the latent factors. It mainly uses two types of likelihood: Gaussian and Poisson, to model the data. The Gaussian likelihood setting focuses on non-negative real-valued data. It usually imposes truncated Gaussian as the prior on each latent factor~\citep{Hinrich2018variational}. There has also been work on imposing a hierarchical structure on the prior using conjugate distributions~\citep{schmidt2009probabilistic}.

\subsection{Bayesian Poisson-Gamma Tensor Factorization (BPTF)}
Gaussian NTF, however, is not good at describing real-world tensor data whose distribution is typically concentrated on \textit{zeros} (used to represent the missing values) and \textit{long-tailed}. Using multi-way publication data as an example, authors typically use small fractions of the vocabulary to write their articles, causing counts of most words to be zero. In their articles, words with larger counts are also less likely to be observed.

The above issues prompted the following work~\citep{schein2015bayesian,Schein:2016:BPT:3045390.3045686} to alternatively use the Poisson log-likelihood to fit the data. It normally yields a better fit since it naturally ignores zero values and can capture the long tail of word counts. In this case, each latent factor follows a Gamma distribution. This distribution imposes the non-negativity constraint on the latent factors. Meanwhile, it can induce sparsity on the factors, which means that most of them become close to zero. This causes the Poisson distributions constructed by these latent factors to become long-tailed. 

For the Poisson distribution, both the mean and the variance are equal to the rate. Some recent work~\citep{hu2015scalable} has also dealt with the case where the variance is much larger than the mean, called the over-dispersion of count data. In this case, the negative binomial distribution, whose variance is strictly larger than the mean, is used to construct the likelihood~\citep{pmlr-v22-zhou12c}. Based on this setting, Hu et al.~\citep{hu2015scalable} proposed a novel online inference algorithm for handling massive tensors. In this paper, we adhere to the Poisson likelihood modelling assumption and show that our proposed framework can still outperform the negative binomial model. 
\subsection{Auxiliary Variable Augmentation}\label{sec:variable_augmentation}
BPTF aims to infer the \textit{joint posterior distribution} of the latent factors given the data and the hyper-parameters. Since deriving this distribution is analytically intractable, the previous work had to resort to approximation techniques such as the MCMC algorithms (e.g. Gibbs sampling) and the variational inference algorithms. The prerequisite step for these algorithms is the auxiliary variable augmentation. It adds a generative layer of latent counts between each observed data instance and its associated Poisson distribution. The additional latent count layer endows the factorization models with the Poisson-Gamma conjugacy. It allows the approximation techniques to be conducted in closed forms. 
\subsection{Deep Learning for Matrix Factorization}
In recent years, various deep learning models have been applied to matrix completion and factorization in recommendation systems\footnote{For a more comprehensive review on this subject, we refer readers to~\citep{Zhang2019survey}.}. They include multi-layer perceptron~\citep{he2017neural,ijcai2017-447}, convolutional networks~\citep{Kim2016CMF,He2018OPN}, auto-encoders~\citep{Sedhain2015AAM,Li2015DCF} and recurrent networks~\citep{HidasiKBT15}. Most of them embedded different combinations of three types of vectors: user, item and their pair-wise interaction. The vectors contain varieties of information regarding the three. Typically, for either user or item vectors, they contain the IDs and ratings specific to the user/item. They can also contain side information. For example, user vectors can contain demographic features while item vectors contain content features. The interaction vectors mostly contain pairs of user and item IDs. The user/item embedding vectors reconstruct the rating matrix via decoding layers~\citep{Sedhain2015AAM} or traditional matrix factorization~\citep{Li2015DCF,ijcai2017-447}. Alternatively, they and their interaction vectors can be fed into the various deep neural networks~\citep{he2017neural,Kim2016CMF,He2018OPN,HidasiKBT15} to predict the ratings. 

As for specifically extending the Poisson-Gamma factor models with neural networks, \citet{yu2019neural} proposed to integrate neural pairwise ranking into the factor models for collaborative filtering on count data. It replaced the likelihood objective of the Poisson factor models with a pairwise ranking function. This function is modelled by a neural network instead of a traditional linear ranking function. The network predicts rankings on pairs of items rated by the same user based on non-linear transformation of the concatenated latent factors between the user and each of the items. The new ranking objective is then regularized by the Gamma priors imposed by the factor models over the latent factors of the users and items. However, this model is user-oriented and thus cannot be directly applied to non-user-oriented cases in matrix and tensor factorization.

Deep learning based matrix completion and factorization have achieved state-of-the-art performance. However, most of them address the ad-hoc problem and are not scalable to tensor factorization. Furthermore, many of them require side information (e.g. user demographics and sessions, item content) to be contained by the user/item feature vectors. This limits their general application. Finally, most of these models tend not to fit count data well. This is because the probability distributions they used for their output layers were either Gaussian or Categorical, instead of the Poisson.

\subsection{Deep Learning for Tensor Factorization}
Generalized from matrices, tensors contain much sparser data and thus require the factorization techniques to capture more complex underlying patterns of the interactions among different entities. Recently, there starts to emerge research work that leverages deep neural networks for constructing non-linear neural factorization models. These models have been shown to generally outperform traditional multilinear factorization models (e.g. CP decompoistion~\citep{kolda2009tensor}) for sparse tensor completion.

\citet{xianntf2019} proposed a neural tensor factorization model which specifically considers the temporal information of the interactions among different entities. This model leverages an LSTM network \citep{hochreiter1997long} to encode the timestamps into corresponding time embeddings. They are then fed into a MLP network, along with the embeddings from the other modes, to perform rating and link prediction for tensors under different applications. However, due to this blackbox LSTM component, this model is only ad-hoc to applications that involve temporal information.

\citet{LiuCoSTCo2019} proposed to use convolutional neural networks (CNNs) \citep{Krizhevskycnn2012} to capture non-linear complex patterns of interactions among entities inside tensors. The model is generalized enough to be applied to any tensor factorization problem but meanwhile lacks dedicated design for handling imbalanced count data. Furthermore, many of the factorization problems may not exhibit (repeatable) local patterns of interactions among entities for the model to work desirably.

\citet{deng2017factorized} proposed the factorized VAE (FVAE) for audience (facial expression) analysis. Its design is based on a specific graphical model for the particular analysis. The model imposes informative priors, factorized over latent audience and time factors, to sample some intermediate latent factors (which correspond to samples drawn from the encoders of FVAE). The intermediate factors then interact with the spatial factors of facial landmarks (which correspond to the decoder) to generate the landmark locations. In comparison, our framework is designed based on a general graphical model without any ad-hoc priors. FVAE infers the posterior distributions of the intermediate latent factors to capture dynamic audience reactions. It ignores the posterior inference for the latent factors from each tensor mode. Our framework performs the latter inference which is a more general case of Bayesian tensor factorization.

\section{Problem Formulation}\label{sec:problem}
Suppose that there is a set of occurred events which can be generally described as ``user $u$ interacted with item $v$ at time $t$''. The counts of their occurrences can be formally represented by a three-way tensor $\boldsymbol{Y}$ of size $|\mathcal{U}| \times |\mathcal{V}| \times |\mathcal{T}|$. The symbols $\mathcal{U}, \mathcal{V}$ and $\mathcal{T}$ respectively denote the sets of users, items and time steps involved in the events. Most of the entries in this tensor are zero (for missing/non-occurred events). The remaining entries record the counts of the events' occurrences. In this paper, we will build a tensor factorization framework which reconstructs $\boldsymbol{Y}$ with multi-dimensional latent factor vectors $\boldsymbol{z}_u, \boldsymbol{z}_v$ and $\boldsymbol{z}_t$, with $u \in \mathcal{U}, v \in \mathcal{V}$ and $t \in \mathcal{T}$. The reconstruction aims to minimize the total difference between each entry $y_{uvt}$ in $\boldsymbol{Y}$ and their predictions $\hat{y}_{uvt}$. We use symbol $\mathcal{X}_{:,:,:}$ to express the set of index triplets corresponding to each non-zero entry in $\boldsymbol{Y}$. Likewise, symbols $\mathcal{X}_{u,:,:}$, $\mathcal{X}_{:,v,:}$ and $\mathcal{X}_{:,:,t}$ denote the sets of index pairs for each non-zero entry specific to user $u$, item $v$ and time $t$ respectively. In general, we use round brackets ($\cdot,...,\cdot$) to denote tuples and angular brackets $\langle\cdot,...,\cdot\rangle$ to denote vectors.

\subsection{Limitations of Auxiliary Variable Augmentation}~\label{sec:variable_augmentation_problem}

A major limitation of the auxiliary variable augmentation resides in the update rules it imposes on the Gamma rate and shape for the latent factors. More specifically, consider $z_{uk}$, the $k$-th component of the latent factor vector for user $u$. In Gibbs sampling, its posterior distribution is the following Gamma distribution:
\begin{eqnarray}
    z_{uk} \sim \text{Gamma}(\alpha_{uk},\beta_{uk})
\label{eqn:gamma_user}
\end{eqnarray}
\begin{eqnarray}
\alpha_{uk} = \alpha+\sum_{(v,t)\in \mathcal{X}_{u,:,:}}c_{uvt,k}
\label{eqn:gamma_user_alpha}
\end{eqnarray}
\begin{eqnarray}
\beta_{uk} = \beta+\sum_{(v,t)\in \mathcal{X}_{u,:,:}}z_{vk}\times z_{tk}\label{eqn:gamma_user_beta}
\end{eqnarray}

In the above equations, $\alpha$ and $\beta$ are the shape and rate of the Gamma prior over $z_{uk}$. Equation~\ref{eqn:gamma_user_alpha} computes the posterior shape $\alpha_{uk}$ using the $k$-th auxiliary latent counts $\{c_{uvt,k}\}_{(v,t)\in \mathcal{X}_{u,:,:}}$. Equation~\ref{eqn:gamma_user_beta} computes the posterior rate $\beta_{uk}$ using pairs of item and time factors $\{(z_{vk}, z_{tk})\}_{(v,t)\in \mathcal{X}_{u,:,:}}$. The auxiliary latent count $c_{uvt,k}$ is drawn alternately with $z_{uk}, z_{vk}$ and $z_{tk}$ from the following Poisson distribution:
\begin{eqnarray}
    c_{uvt,k} \sim \text{Poisson}(z_{uk}\times z_{vk}\times z_{tk})\label{eqn:latent_count}
\end{eqnarray}
Variational inference algorithms have similar update procedures. They require parameter updates for the additional multinomial distributions over the auxiliary latent counts\footnote{For a more detailed mathematical description, we refer readers to~\citep{Gopalan2015SRH}.}.

In Equations \ref{eqn:gamma_user_alpha} and \ref{eqn:gamma_user_beta}, the latent variables (i.e. $\{(z_{vk}, z_{tk}, c_{uvt,k})\}_{(v,t)\in \mathcal{X}_{u,:,:}}$) used to update the rate and shape parameters correspond only to the data of user $u$. This means that these updates fail to utilize the data from other similar users for calibration. Moreover, the update formulas shown in the two equations were derived based on the Poisson-Gamma conjugacy. They do not necessarily reflect the underlying mappings between the posterior parameters and the latent variables. We want the update rules to have more flexibility to capture possibly complex mappings.

\section{Proposed Framework: VAE-BPTF}
We now describe our framework: \textbf{V}ariational \textbf{A}uto\textbf{E}ncoder based \textbf{B}ayesian \textbf{P}oisson-gamma \textbf{T}ensor \textbf{F}actorization (VAE-BPTF). To make the description succinct, we consider the three-way tensor $\boldsymbol{Y}$ and its factorized latent vectors $\boldsymbol{z}_u, \boldsymbol{z}_v$ and $\boldsymbol{z}_t$ introduced in Section~\ref{sec:problem}. 

The central part of VAE-BPTF is the inference of the posterior Gamma distributions for the latent factors. VAE-BPTF adopts the mean-field variational inference. It assumes that the latent factors are independent and estimates their posterior Gamma rates and shapes. It does this by maximizing the evidence lower bound (ELBO) $\text{Q}$ for the Poisson likelihood over $\boldsymbol{Y}$ which is expressed as follows:
\begin{equation}
\begin{split}
\text{Q}=\sum_{(u,v,t) \in \mathcal{X}_{:,:,:}}&\bigg(y_{uvt}\ln\lambda_{uvt}-\lambda_{uvt}\bigg)-\sum_{k\in \mathcal{K}}\Bigg[\sum_{u\in \mathcal{U}}\text{KL}\bigg(\text{Gamma}\big(\alpha_{uk},\beta_{uk}\big)||\\&\text{Gamma}\big(\alpha,\beta\big)\bigg)-\sum_{v\in \mathcal{V}}\text{KL}\bigg(\text{Gamma}\big(\alpha_{vk},\beta_{vk}\big)||\text{Gamma}\big(\alpha,\beta\big)\bigg)-\\&~~~~~~~~~~~~~~~~~~~~~~~~~~\sum_{t\in \mathcal{T}}\text{KL}\bigg(\text{Gamma}\big(\alpha_{tk},\beta_{tk}\big)||\text{Gamma}\big(\alpha,\beta\big)\bigg)\Bigg]
\end{split}~\label{eqn:elbo_our_model}
\end{equation}
In Equation~\ref{eqn:elbo_our_model}, the Poisson rate $\lambda_{uvt}$ is specific to the entry $y_{uvt}$ and is computed according to the CP decompoistion~\citep{kolda2009tensor} as follows:
\begin{equation}
    \lambda_{uvt} = \sum_{k\in \mathcal{K}}z_{uk}\times z_{vk}\times z_{tk}
\end{equation}
The KL divergence between the posterior Gamma (e.g. parameterized by $\alpha_{uk}, \beta_{uk}$) and prior Gamma distributions is calculated as follows:
\begin{align}
\small
\begin{split}
\text{KL}\bigg(&\text{Gamma}\big(\alpha_{uk},\beta_{uk}\big)||\text{Gamma}\big(\alpha,\beta\big)\bigg)=\\&(\alpha_{uk}-\alpha)\psi(\alpha_{uk})-\log\Gamma(\alpha_{uk})+\log\Gamma(\alpha)+\alpha(\log\beta_{uk}-\log\beta)+\alpha_{uk}\frac{\beta-\beta_{uk}}{\beta_{uk}}
\end{split}~\label{eqn:kl_divergence}
\end{align}
where $\Gamma(\cdot)$ and $\psi(\cdot)$ are respectively the gamma and digamma functions.

Classical BPTF relies on the auxiliary variable augmentation. As discussed in Section~\ref{sec:variable_augmentation_problem}, the augmentation enables closed-form posterior updates on the Gamma rates and shapes for latent factors. However, the update formulas are limited in their flexibility and expressive power. Furthermore, the updates fail to share information across users (as well as items and time steps). They are susceptible to the possible sparsity in the individuals' data. This leads to an unreliable estimation of the posterior Gamma rates and shapes. 

Combining BPTF with VAE can solve both problems. VAEs gain their expressive power via activation functions and depths. Meanwhile, the weights between layers can map similar inputs to similar outputs. This exerts a smoothing effect on noisy sparse data information of individuals. 

\subsection{Basic Framework Structure}
\begin{figure*}[t]
\begin{subfigure}[b]{0.3\columnwidth}
  \centering
\includegraphics[width=1.5in]{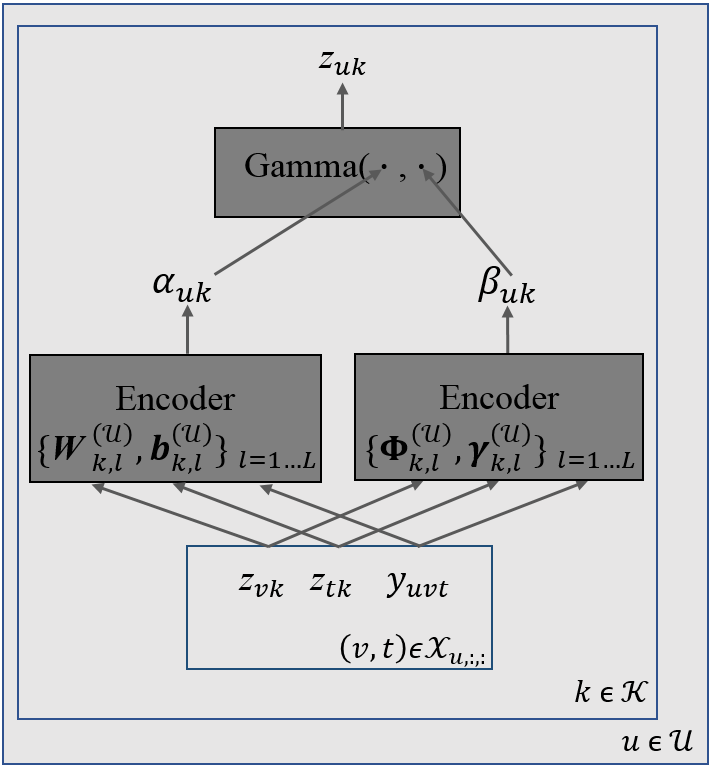}
  \caption{VAE-BPTF's $k$-th encoders specific to mode ``User'' that compute the posterior shape $\alpha_{uk}$ and rate $\beta_{uk}$ for each user $u \in \mathcal{U}$.}\label{fig:vae_bptf_architecture1}
  \end{subfigure}\hspace{0.3cm}
  \begin{subfigure}[b]{0.3\columnwidth}
  \centering
\includegraphics[width=1.25in]{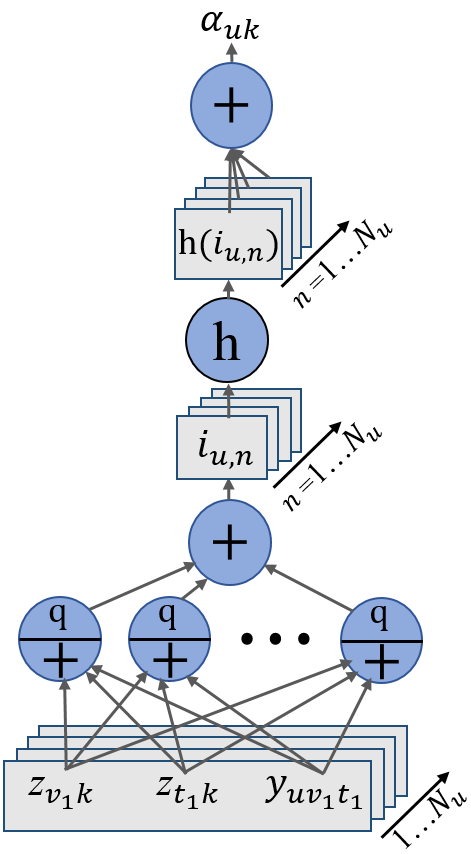}
  \caption{Inside the $k$-th encoder that computes $\alpha_{uk}$ for $u \in \mathcal{U}$. $N_u$ inputs pass through a series of non-linear transformation before summed up.}\label{fig:vae_bptf_architecture2}
  \end{subfigure}\hspace{0.3cm}
  \begin{subfigure}[b]{0.3\columnwidth}
  \centering
\includegraphics[width=1.25in]{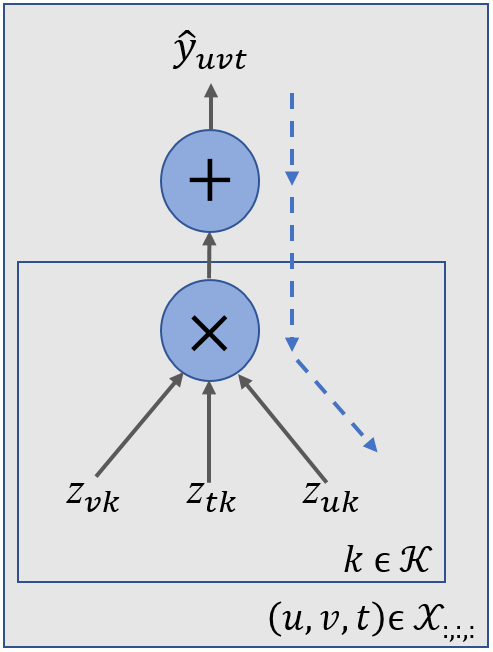}
  \caption{VAE-BPTF's decoder that reconstructs each data point $y_{uvt}$. Blue dashed lines show the chain of gradients with respect to parameters of the $k$-th encoders specific to mode ``User''.}\label{fig:vae_bptf_architecture3}
  \end{subfigure}

\caption{The architecture of the VAE-BPTF framework.}\label{fig:vae_bptf_architecture}
\end{figure*}

Figure~\ref{fig:vae_bptf_architecture} shows the architecture of the VAE-BPTF framework. It inherits the \textit{encoder-decoder} structure from the auto-encoders. Figure~\ref{fig:vae_bptf_architecture3} shows the decoder part of the framework. It reconstructs each observed data $y_{uvt}$ using the CP decomposition. Figure~\ref{fig:vae_bptf_architecture1} displays the encoder part. There are $|\mathcal{K}|$ encoders dedicated to computing either the posterior shapes or rates specific to each mode of the tensor $\boldsymbol{Y}$. In this case, the total number of encoder networks in VAE-BPTF is $\text{3}\times\text{2}\times|\mathcal{K}|$ as there are three modes in the tensor and two parameters in the Gamma distribution. The $k$-th encoders that respectively compute the posterior shape $\alpha_{uk}$ and the rate $\beta_{uk}$ replace the conjugate updates in Equations~\ref{eqn:gamma_user_alpha} and~\ref{eqn:gamma_user_beta} as follows:
\begin{eqnarray}
\begin{split}
\alpha_{uk}=\sum_{(v,t)\in \mathcal{X}_{u,:,:}}\text{h}(\boldsymbol{f}^{(\mathcal{U})}_{uvt,k,L}\boldsymbol{w}^{(\mathcal{U})}_{k,L+1}+b^{(\mathcal{U})}_{k,L+1})\\
\beta_{uk}=\sum_{(v,t)\in \mathcal{X}_{u,:,:}}\text{h}(\boldsymbol{g}^{(\mathcal{U})}_{uvt,k,L}\boldsymbol{\phi}^{(\mathcal{U})}_{k,L+1}+\gamma^{(\mathcal{U})}_{k,L+1})
\end{split}~\label{eqn:encoder_fun}
\end{eqnarray}
For each pair $(v,t)\in \mathcal{X}_{u,:,:}$, we have the following recursive equations:
\begin{flalign}
\begin{split}
\boldsymbol{f}^{(\mathcal{U})}_{uvt,k,L}=\text{q}(\boldsymbol{f}^{(\mathcal{U})}_{uvt,k,L-1}\boldsymbol{W}^{(\mathcal{U})}_{k,L}+\boldsymbol{b}^{(\mathcal{U})}_{k,L})\\
\ldots\hspace{2.5cm}\\
\boldsymbol{f}^{(\mathcal{U})}_{uvt,k,1}=\text{q}(\boldsymbol{f}^{(\mathcal{U})}_{uvt,k,0}\boldsymbol{W}^{(\mathcal{U})}_{k,1}+\boldsymbol{b}^{(\mathcal{U})}_{k,1})\\\\
\boldsymbol{g}^{(\mathcal{U})}_{uvt,k,L}=\text{q}(\boldsymbol{g}^{(\mathcal{U})}_{uvt,k,L-1}\boldsymbol{\Phi}^{(\mathcal{U})}_{k,L}+\boldsymbol{\gamma}^{(\mathcal{U})}_{k,L})\\
\ldots\hspace{2.5cm}\\
\boldsymbol{g}^{(\mathcal{U})}_{uvt,k,1}=\text{q}(\boldsymbol{g}^{(\mathcal{U})}_{uvt,k,0}\boldsymbol{\Phi}^{(\mathcal{U})}_{k,1}+\boldsymbol{\gamma}^{(\mathcal{U})}_{k,1})\\\\
\boldsymbol{f}^{(\mathcal{U})}_{uvt,k,0}=\boldsymbol{g}^{(\mathcal{U})}_{uvt,k,0}=\langle z_{vk},z_{tk},y_{uvt}\rangle\\
\end{split}~\label{eqn:recursive_fun}
\end{flalign}
In Equation~\ref{eqn:encoder_fun}, the set of input vectors fed to the two encoders' output layers, indexed by ($L+1$), are $\{ \boldsymbol{f}^{(\mathcal{U})}_{uvt,k,L}, \boldsymbol{g}^{(\mathcal{U})}_{uvt,k,L}\}_{(v,t)\in \mathcal{X}_{u,:,:}}$. 
The weight vectors of these output layers are $\boldsymbol{w}^{(\mathcal{U})}_{k,L+1}$ and $\boldsymbol{\phi}^{(\mathcal{U})}_{k,L+1}$ respectively. The scalars $b^{(\mathcal{U})}_{k, L+1}$ and $\gamma^{(\mathcal{U})}_{k, L+1}$ are the respective biases.

\begin{figure}[t]
\centering
\includegraphics[width=3in]{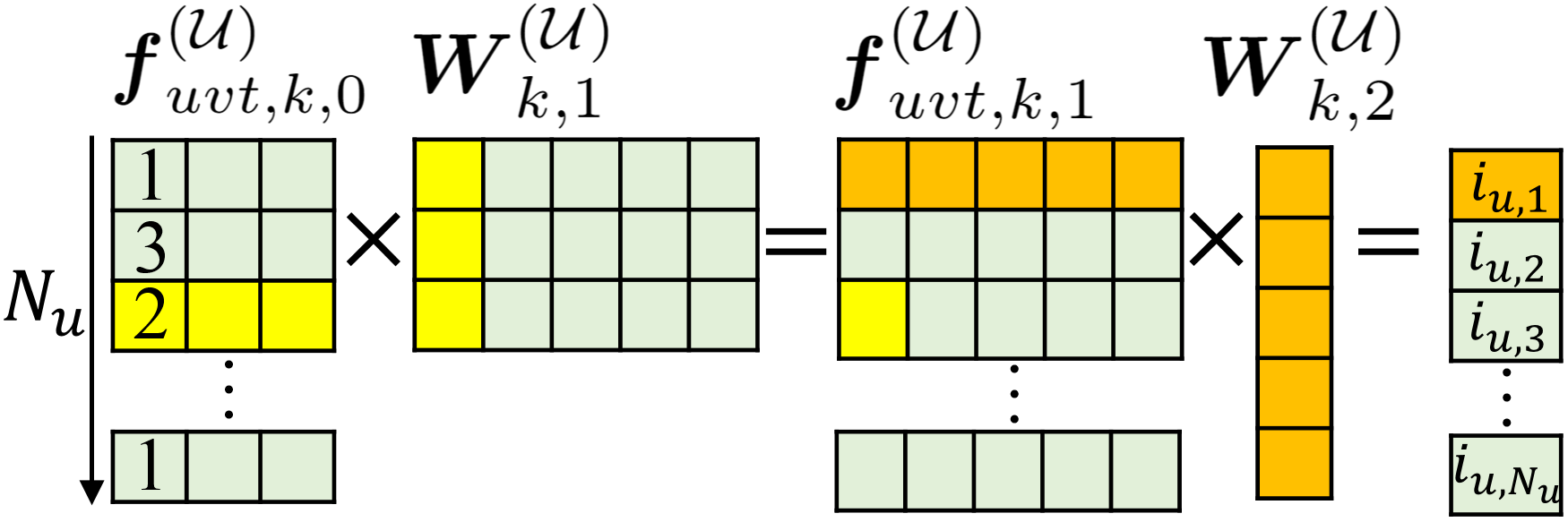}
  \caption{Example of the linear transformation performed by the encoder's weights. The input matrix first multiplies the hidden layer weight matrix. Then, the resulting matrix multiplies the output layer weight vector to obtain the instance-wise contributions.}\label{fig:reweight_scheme_a}
\end{figure}

Equation~\ref{eqn:recursive_fun} shows how $ \boldsymbol{f}^{(\mathcal{U})}_{uvt,k,L}$ and $\boldsymbol{g}^{(\mathcal{U})}_{uvt,k,L}$ are computed recursively from their corresponding input layers $\boldsymbol{f}^{(\mathcal{U})}_{uvt,k,0}$ and $\boldsymbol{g}^{(\mathcal{U})}_{uvt,k,0}$. Both the input layers take in each vector associated with user $u$: $\{\langle z_{vk},z_{tk},y_{uvt}\rangle\}_{(v,t)\in \mathcal{X}_{u,:,:}}$. To compute $\boldsymbol{f}^{(\mathcal{U})}_{uvt,k,L}$, these vectors are fed into $L$ MLP hidden layers. The $l$-th ($1\leq l \leq L$) hidden layer has a weight matrix $\boldsymbol{W}^{(\mathcal{U})}_{k,l}$ shared across all the users. Its number of rows equals the number of neurons in the ($l$-1)-th hidden layer if $l\geq 2$ or otherwise, the number of input features. The number of columns equals the number of neurons in the $l$-th hidden layer. Likewise, the encoder that computes $\boldsymbol{g}^{(\mathcal{U})}_{uvt,k,L}$ has $L$ hidden layers with each layer having a weight matrix $\boldsymbol{\Phi}^{(\mathcal{U})}_{k,l}$ ($1\leq l \leq L$). 

Figure~\ref{fig:vae_bptf_architecture2} illustrates the inner structure of the $k$-th encoder with one hidden layer that computes $\alpha_{uk}$ based on Equations~\ref{eqn:encoder_fun} and \ref{eqn:recursive_fun}. Figure~\ref{fig:reweight_scheme_a} further illustrates the linear transformation performed by the weights\footnote{For simplicity, we omitted the activation functions and the bias terms in between.} of the hidden and output layers. The inputs are organized into \textit{user-specific batches}. The size of the batch for user $u$ is $N_u=|\mathcal{X}_{u,:,:}|$. The hidden layer in this case has five neurons with a 3$\times$5 weight matrix $\boldsymbol{W}^{(\mathcal{U})}_{k,1}$. The output layer maps the hidden neuron outputs linearly with its 5$\times$1 weight vector $\boldsymbol{w}^{(\mathcal{U})}_{k,2}$. This yields a set of scalars $\{i_{u,n}\}_{1 \leq n \leq N_{u}}$ for the batch of user $u$. The scalar $i_{u,n}$ indicates the importance of the $n$-th input vector in the batch in predicting $\alpha_{uk}$.    

\begin{figure}[t]
\centering
\includegraphics[width=1.75in]{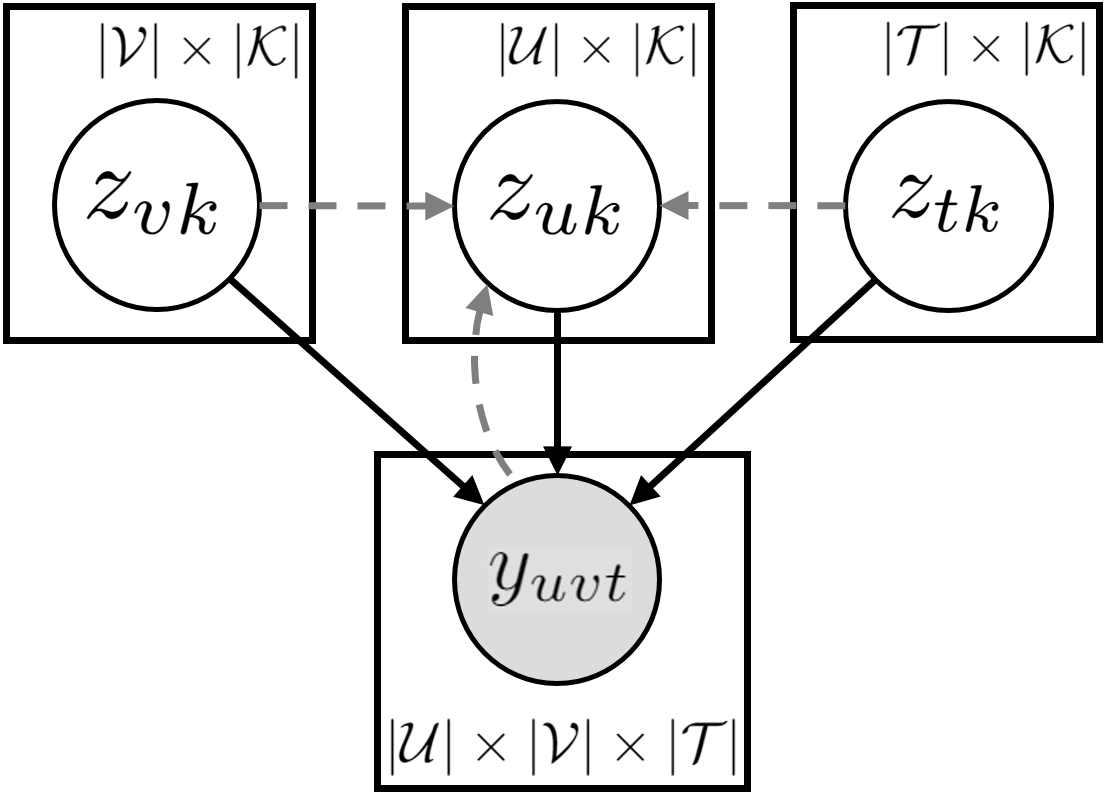}
  \caption{VAE-BPTF depicted as a graphical model. The solid lines show the data generation. The dashed lines show that $z_{uk}$ is inferred using the other latent factors and the generated data.}
\label{fig:generative_inference_process}
\end{figure}

Essentially, Equations~\ref{eqn:encoder_fun} and~\ref{eqn:recursive_fun} embody the inference of the following posterior distribution:\\
\begin{eqnarray}
\begin{split}
p(z_{uk}|\{(z_{vk}, z_{tk}, y_{uvt})\}_{(v,t)\in \mathcal{X}_{u,:,:}}) \propto p(z_{uk}) \times \prod_{(v,t)\in \mathcal{X}_{u,:,:}} p(y_{uvt}|z_{uk}, z_{vk}, z_{tk})
\end{split}~\label{eqn:posterior_z}
\end{eqnarray}The above equation can be further described by the generative and inference processes shown in Figure~\ref{fig:generative_inference_process}. It shows that data instances were generated by the latent factors from each mode. In reverse, the posterior distribution of a latent factor under a particular mode is inferred based on the latent factors under the other modes and the data instances they together generated\footnote{For simplicity, we omitted the prior shape $\alpha$ and rate $\beta$ in Equation~\ref{eqn:posterior_z} and in Figure~\ref{fig:generative_inference_process}. They are not directly used to compute $\alpha_{uk}$ and $\beta_{uk}$ in Equations~\ref{eqn:encoder_fun} and~\ref{eqn:recursive_fun}. Instead, they are leveraged by the KL regularization in Equation~\ref{eqn:elbo_our_model}.}.

\subsection{Sparsity-inducing \& Numerically Stable Activation Functions}
The activation functions $\text{h}(\cdot)$ and $\text{q}(\cdot)$ provide the non-linear transformation for the output and hidden layers respectively. In terms of $\text{h}(\cdot)$, it must not violate the non-negativity constraint on $\alpha_{uk}$ and $\beta_{uk}$. Possible choices of h$(\cdot)$ include the sigmoid function, the softplus function and the rectified linear (ReLU) function. 

To choose $\text{h}(\cdot)$, we consider a trade-off: the \textit{sparsity-inducing} ability of h$(\cdot)$ against the \textit{numerical stability} for which $\alpha_{uk} > 0$ and $\beta_{uk} > 0$. For ReLU, i.e. h$(i)$=max$(0,i)$, we found in the experiments that it could induce sparsity across the inputs. More specifically, it transformed inputs deemed unimportant in predicting $\alpha_{uk}$ or $\beta_{uk}$ into zero. Meanwhile, it kept the values of important inputs. On the other hand, however, we found that ReLU almost certainly failed to yield positive shape and rate for some individuals. They became negative usually after a few training iterations under random initialization on latent factors and network parameters.

In comparison, we found that the softplus function, i.e. $\text{h}(i)$=ln$(1+$exp$(i))$, never led to negative shapes and rates during training. However, its sparsity-inducing ability is weaker than ReLU. This is also the case for the sigmoid function, i.e. $\text{h}(i)$=$1/(1+$exp$(-i))$. Empirically, we found that our framework using the softplus function for both $\text{h}(\cdot)$ and $\text{q}(\cdot)$ overall yielded the best performance compared to other combinations of activation functions\footnote{The combinations include either softplus or sigmoid for $\text{h}(\cdot)$, and either them or ReLU for $\text{q}(\cdot)$.}. 

\subsection{Handling Imbalanced Data}
\begin{figure}[t]
\centering
\includegraphics[width=2.5in]{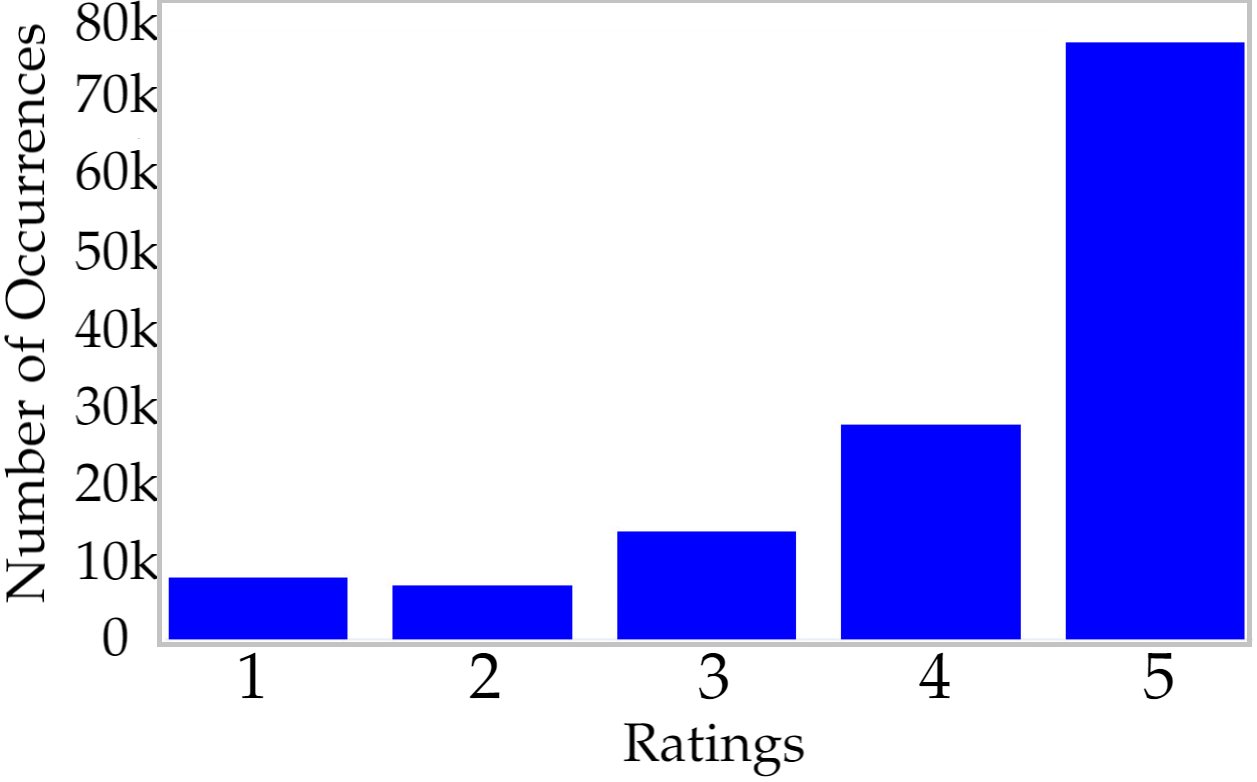}
  \caption{Counts of different ratings across Amazon Prime videos. This data is highly imbalanced.}~\label{fig:imbalanced_amazon_data}
\end{figure}

Imbalanced data values are prevalent in the tensors of discrete data. For example, in text analysis, small count values are more likely to be found in an author-word-year tensor as each author tends to write most words few times in an article. In recommender systems, a user-item-time tensor is more likely to contain high ratings as users tend to give such ratings to items that they like but unwilling to rate those they dislike. Figure~\ref{fig:imbalanced_amazon_data} illustrates the imbalanced ratings on Amazon Prime videos. The imbalance problem has posed a significant challenge to Poisson tensor factorization approaches for making reliable predictions. 

In recommender systems, most matrix factorization models dedicate a global variable $\mu$ to account for the users' average rating. This variable mitigates the imbalance effect by removing the \textit{population bias} underneath. It allows the models to fit $(\boldsymbol{Y} - \mu\boldsymbol{\mathbbm{1}})$ rather than the imbalanced $\boldsymbol{Y}$\footnote{The symbol $\boldsymbol{\mathbbm{1}}$ denotes a matrix of the same size as $\boldsymbol{Y}$ and contains all ones.}. Latent factors are now learned solely based on the \textit{personal bias} information contained by the deviations between the ratings and the average. Poisson factorization models are not compatible with a mean variable $\mu$ as $(\boldsymbol{Y} - \mu\boldsymbol{\mathbbm{1}})$ can contain negative entries the models cannot factorize. However, without a proper way of handling the population bias, the training of the models will end up just learning the average rating.

\begin{figure}[t]
  \centering
\includegraphics[width=2.75in]{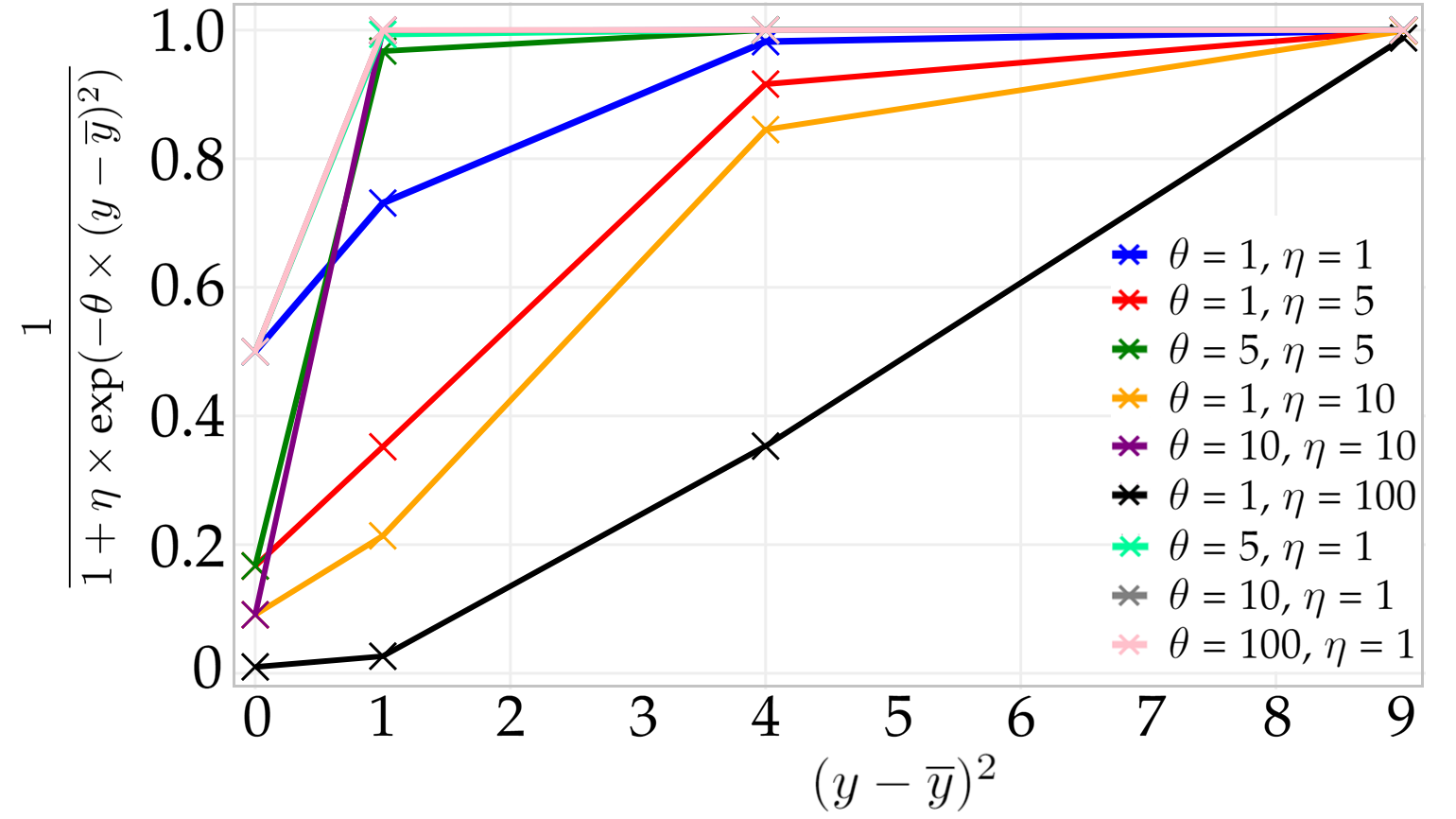}
  \caption{The reweighting function with different parameter values varies across the squared distance.}~\label{fig:reweight_fun}
\end{figure}

To enable VAE-BPTF to address this issue, we need to reweigh activation results from its encoders' output layers. Our design of the reweighting scheme is inspired by two lines of research. \citet{ahn2008new} has proposed a pairwise metric to measure the importance of every pair of ratings on the same items from two users to their similarity. It states that when two users' ratings on the same item are close to the average rating of the item, the agreement between the two ratings might not provide much information about the similarity. In contrast, if the two ratings are close and also far from the average rating, it indicates a stronger similarity of two users in their tastes. The above idea reflects that ratings farther from the average tends to yield more information about the characteristics of users. Our reweighting scheme is partially inspired by this idea but focuses on measuring the importance of a single rating to each entity associated with it.

\citet{hu2008collaborative} have proposed a reweighting scheme for matrix factorization on implicit feedback data, which can be extended to tensors. Their scheme is integrated into the objective function of maximum a posterior inference for the latent factors of the entities (e.g. users):
    \begin{equation}
        \log p\big(\boldsymbol{z}_u|\{\boldsymbol{z}_{v}, \boldsymbol{z}_{t}, y_{uvt}\}_{(v,t)\in \mathcal{X}_{u,:,:}}\big)\propto \log p(\boldsymbol{z}_{u}) + \sum_{(v,t)\in \mathcal{X}_{u,:,:}}\Delta_{uvt}\log p(y_{uvt}|\boldsymbol{z}_{u},\boldsymbol{z}_{v},\boldsymbol{z}_{t})
    \end{equation}
    where $\Delta_{uvt}$ is the reweighting term specific to the entry $y_{uvt}$. The authors dealt with binary data (e.g. click data) and therefore proposed a strictly increasing reweighting function:
        \begin{equation}
        \Delta_{uvt}=1+\theta y_{uvt}
    \end{equation}
    where $\theta$ is a positive hyper-parameter. According to the authors, this function down-weights the unobserved values in the binary matrix and helps amplify the evidence for positive preference. Since we deal with explicit feedback data (i.e. $y_{uvt}>0$), the reweighting function needs to be designed to suppress the average or the most frequent observed value and amplify the evidence for both positive and negative preference relative to such a value.

Motivated by the above work, our reweighting scheme is based on the deviation of each data value from the most frequent count value $\bar{y}$. The smaller the deviation, the smaller the weight becomes. Thus, common data values will have less influence in predicting the posterior parameters. The reweighting function adopted by VAE-BPTF is thus:

\begin{eqnarray}
\begin{split}
\Delta(y,\bar{y}) = \frac{1}{1+\eta\times \text{exp}(-\theta\times (y - \overline{y})^2)}
\end{split}~\label{eqn:reweight_fun}
\end{eqnarray}
In Equation~\ref{eqn:reweight_fun}, the parameters $\theta > 0$ and $\eta > 0$ is the slope and the intercept of the function $\Delta(y,\bar{y})$. Figure~\ref{fig:reweight_fun} shows the values of $\Delta(y,\bar{y})$ over $(y-\bar{y})^2$ for count data under different values for $\theta$ and $\eta$. It can be seen that $\theta$ is much more sensitive to the change in $(y-\bar{y})^2$ than $\eta$ for driving $\Delta(y,\bar{y})$ towards 1. 

\begin{figure}[t]
\centering
\includegraphics[width=3in]{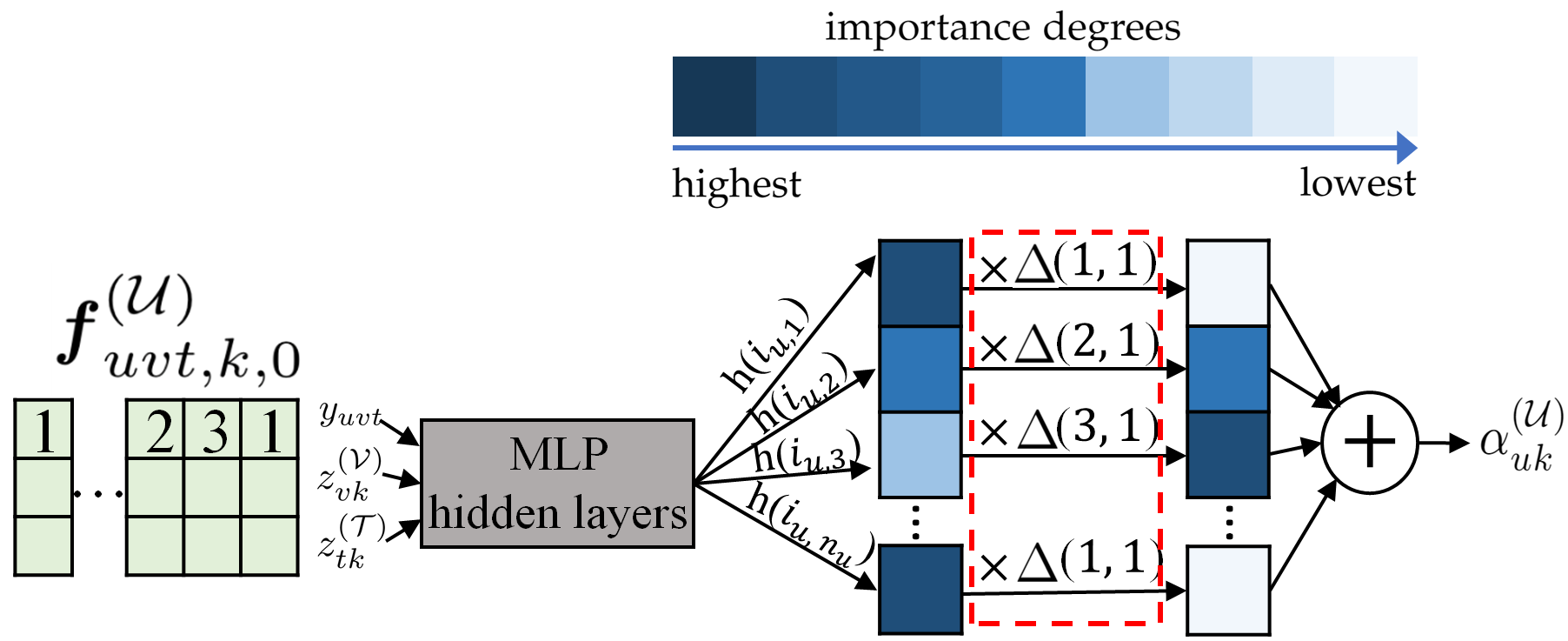}
  \caption{Example of how the reweighting function is integrated into VAE-BPTF's encoders. Here, the previously important contributions, indicated by the dark colours, are penalized to become unimportant due to their associated data value being close to the most frequent value.}
  \label{fig:reweight_scheme_b}
  \end{figure}

We integrate the reweighting function with VAE-BPTF by changing Equation~\ref{eqn:encoder_fun} as follows:

\begin{eqnarray}
\begin{split}
\alpha_{uk}=\sum_{(v,t)\in \mathcal{X}_{u,:,:}}\Delta(y_{uvt}, \bar{y}_{uvt})\times\text{h}(\boldsymbol{w}^{(\mathcal{U})}_{k,L+1}\boldsymbol{f}^{(\mathcal{U})}_{uvt,k,L}+b^{(\mathcal{U})}_{k,L+1})\\
\beta_{uk}=\sum_{(v,t)\in \mathcal{X}_{u,:,:}}\Delta(y_{uvt}, \bar{y}_{uvt})\times\text{h}(\boldsymbol{\phi}^{(\mathcal{U})}_{k,L+1}\boldsymbol{g}^{(\mathcal{U})}_{uvt,k,L}+\gamma^{(\mathcal{U})}_{k,L+1})
\end{split}~\label{eqn:encoder_fun1}
\end{eqnarray}
Figure~\ref{fig:reweight_scheme_b} illustrates reweighting the activation function values specific to user $u$. The majority of the count values, i.e. count 1 in this case, have the least weight for their activation values (or equivalently, the least importance in predicting $\alpha_{uk}$). Meanwhile, count 3 has a larger weight as it is farther from count 1. Depending on the extent of the imbalance, the reweighting function can be made different by varying $\theta$ and $\eta$. 

\section{Gradient Reparametrization}
For any framework that exerts variational inference, including VAE-BPTF, a key problem is to compute gradients of randomly generated latent variables with respect to their posterior parameters. This is not directly feasible due to the stochastic nature of the variables with Markov Chain Monte Carlo (MCMC) methods. A popular solution is called the reparameterization trick. It transforms the posterior distributions of the variables into some differentiable functions in terms of the posterior parameters. These functions make the posterior parameters independent of the stochasticity by using auxiliary random noise to account for it instead. 

A few distributions (e.g. the Normal distribution) have been proved to have closed-form reparameterized functions. However, many other distributions, including the Gamma distribution, fail to yield analytic forms for the functions. Instead, numerical approximations of the functions have been leveraged in recent work~\citep{figurnov2018implicit,jankowiak2018pathwise,knowles2015stochastic}. 

In this paper, we employ the approximation techniques from \citep{jankowiak2018pathwise}. More specifically, we first specify the gradients of the ELBO function $\text{Q}$ with respect to the posterior rate, e.g. $\beta_{uk}$, as follows:
\begin{eqnarray}
\begin{split}
\nabla_{\beta_{uk}} \text{Q} = \frac{\partial \text{LL}}{\partial z_{uk}} \times \frac{\partial z_{uk}}{\partial \beta_{uk}} - \frac{\partial \text{KL}(\beta_{uk})}{\partial \beta_{uk}}  
\end{split}~\label{eqn:gradient_Q_beta}
\end{eqnarray} 
In Equation~\ref{eqn:gradient_Q_beta}, the symbol $\text{LL}$ denotes the Poisson Log-Likelihood and the term $\frac{\partial \text{LL}}{\partial z_{uk}}$ is calculated as follows: 
\begin{eqnarray}
\frac{\partial \text{LL}}{\partial z_{uk}}=\sum_{(v,t) \in \mathcal{X}_{u,:,:}}&\bigg(\frac{y_{uvt}z_{vk}z_{tk}}{\sum_{k'\in \mathcal{K}}z_{uk'}z_{vk'}z_{tk'}}-z_{vk}z_{tk}\bigg)~\label{eqn:gradient_Q_z}
\end{eqnarray}
The term $\frac{\partial \text{KL}(\beta_{uk})}{\partial \beta_{uk}}$ can be calculated from Equation~\ref{eqn:kl_divergence} as follows:
\begin{eqnarray}
\frac{\partial \text{KL}(\beta_{uk})}{\partial \beta_{uk}}=\alpha\beta^{-1}_{uk}-\beta\alpha_{uk}\beta^{-2}_{uk}~\label{eqn:gradient_kl_beta}
\end{eqnarray}
To calculate the term $\frac{\partial z_{uk}}{\partial \beta_{uk}}$, the scaling property of the Gamma distribution is utilized. More precisely, $z_{uk} \sim \text{Gamma}(\alpha_{uk},\beta_{uk})$ is the same as $ \frac{\beta\times z_{uk}}{\beta} \sim \text{Gamma}(\alpha_{uk},\beta_{uk})$, and as $(\beta\times z_{uk}) \sim \text{Gamma}(\alpha_{uk},1)$. Using an auxiliary variable $\epsilon_{uk} \sim \text{Gamma}(\alpha_{uk},1)$, the term $\frac{\partial z_{uk}}{\partial \beta_{uk}}$ can be calculated as follows:
\begin{eqnarray}
\frac{\partial z_{uk}}{\partial \beta_{uk}}=\frac{\partial (\epsilon_{uk}/\beta_{uk})}{\partial \beta_{uk}} = -\beta_{uk}^{-2}\times \epsilon_{uk}~\label{eqn:gradient_z_beta}
\end{eqnarray} Likewise, to compute the gradient $\nabla_{\alpha_{uk}} \text{Q}$, the terms $\frac{\partial \text{KL}(\alpha_{uk})}{\partial \alpha_{uk}}$ and $\frac{\partial z_{uk}}{\partial \alpha_{uk}}$ need to be calculated. The former term can be calculated from Equation~\ref{eqn:kl_divergence} as follows:
\begin{eqnarray}
\frac{\partial \text{KL}(\alpha_{uk})}{\partial \alpha_{uk}}=\alpha_{uk}\psi'(\alpha_{uk})-\alpha\psi'(\alpha_{uk})+\beta\beta^{-1}_{uk}-1~\label{eqn:gradient_kl_alpha}
\end{eqnarray}
where $\psi'(\alpha_{uk})$ is the trigamma function. Both the digamma and trigamma functions can be readily computed using any major numerical computation software. 

For the latter term, we need to calculate it as:
\begin{eqnarray}
\frac{\partial z_{uk}}{\partial \alpha_{uk}}=\beta_{uk}^{-1}\times\frac{\partial \epsilon_{uk}}{\partial \alpha_{uk}}~\label{eqn:gradient_z_alpha}
\end{eqnarray} Given that $\epsilon_{uk}$ follows a standard Gamma distribution, computing $\frac{\partial \epsilon_{uk}}{\partial \alpha_{uk}}$ involves the following reparameterization:  \begin{eqnarray}
\begin{split}
\frac{\partial \epsilon_{uk}}{\partial \alpha_{uk}} = \frac{\partial \text{P}(\epsilon_{uk};\alpha_{uk})}{\partial \alpha_{uk}} / \frac{\partial \text{P}(\epsilon_{uk};\alpha_{uk})}{\partial \epsilon_{uk}}\\
= \frac{\partial \text{P}(\epsilon_{uk};\alpha_{uk})}{\partial \alpha_{uk}} / p(\epsilon_{uk};\alpha_{uk})
\end{split}~\label{eqn:gamma_cdf_pdf}
\end{eqnarray} where $\text{P}(\epsilon_{uk};\alpha_{uk})$ is the cumulative distribution function (CDF) of the standard Gamma distribution and~$p(\epsilon_{uk};\alpha_{uk})$ is its probability density function (PDF). For Gamma distributed latent variable $\epsilon_{uk}$, its PDF has a closed form. However, its CDF does not yield an analytically tractable derivative with respect to $\alpha_{uk}$ in Equation~\ref{eqn:gamma_cdf_pdf}. \citet{jankowiak2018pathwise} proposed the following approximation of the derivative\footnote{For more details about the exact formulas of \text{TE}$(\epsilon_{uk};\alpha_{uk})$ and $\text{R}\big(\text{log}(\frac{\epsilon_{uk}}{\alpha_{uk}}), \text{log}(\alpha_{uk})\big)$, and their derivation, we refer readers to the supplementary materials of~\citep{jankowiak2018pathwise}.}: 
\begin{eqnarray}
\frac{\partial \text{P}(\epsilon_{uk};\alpha_{uk})}{\partial \alpha_{uk}}= 
\begin{cases}
    \frac{\partial ~ \text{TE}(\epsilon_{uk};\alpha_{uk})}{\partial \alpha_{uk}},& \epsilon_{uk} \textless 0.8\\
    \frac{\partial}{\partial\alpha_{uk}}\bigg(\sqrt{\frac{\epsilon_{uk}}{\alpha_{uk}}}\bigg),              & \epsilon_{uk} \textgreater 8\\
    \frac{\partial~\text{exp}\bigg[\text{R}\big(\text{log}(\frac{\epsilon_{uk}}{\alpha_{uk}}), \text{log}(\alpha_{uk})\big)\bigg]}{\partial\alpha_{uk}},& \text{otherwise}\\
\end{cases}~\label{eqn:gamma_cdf_approximate}
\end{eqnarray}
In the above equation, \text{TE}$(\epsilon_{uk};\alpha_{uk})$ is a Taylor series expansion of $\text{P}(\epsilon_{uk};\alpha_{uk})$. Meanwhile, $\text{R}\big(\text{log}(\frac{\epsilon_{uk}}{\alpha_{uk}}), \text{log}(\alpha_{uk})\big)$ is a rational polynomial function of orders up to 2 and 3 in coordinates $\text{log}(\frac{\epsilon_{uk}}{\alpha_{uk}})$ and $\text{log}(\alpha_{uk})$ respectively. 



\begin{figure*}[t]
  \begin{algorithm}[H]
   \caption{Inference Scheme of VAE-BPTF}
\small
{Initialize} encoder output layer parameters under each mode $\mathcal{S}=\mathcal{U},\mathcal{V}~\text{or} ~\mathcal{T}$ and $k \in \mathcal{K}$: $\{\boldsymbol{w}^{(\mathcal{S})}_{k,L+1},b^{(\mathcal{S})}_{k,L+1}\}$ and $\{\boldsymbol{\phi}^{(\mathcal{S})}_{k,L+1}, \gamma^{(\mathcal{S})}_{k,L+1}\}$ by sampling each element from $\mathcal{N}(0,\sigma^{2})$ and  $\mathcal{N}(0.1,\sigma^{2})$ respectively;\DontPrintSemicolon\;
{Initialize} encoder hidden layer parameters under each mode: $\{\boldsymbol{W}^{(\mathcal{S})}_{k,l},\boldsymbol{b}^{(\mathcal{S})}_{k,l}, \boldsymbol{\Phi}^{(\mathcal{S})}_{k,l}, \boldsymbol{\gamma}^{(\mathcal{S})}_{k,l}\}^{\mathcal{S}=\mathcal{U},\mathcal{V}~\text{or} ~\mathcal{T}}_{k \in \mathcal{K},~l=1~\text{to}~L}$ by sampling each of their elements from $\mathcal{N}(0,1)$;\\ 
{Initialize} latent factors under each mode: $\{z_{sk}\}_{k \in \mathcal{K}, s \in \mathcal{U},\mathcal{V}~\text{or}~ \mathcal{T}}$ by sampling each of them from Gamma($\alpha$,~$\beta$);\;
Construct input batches for the $k$-th ($k \in \mathcal{K}$) encoders for $\mathcal{U}, \mathcal{V}$ and $\mathcal{T}$: ~~~~~~~~~~~~~~~~~~~~~~~~~~~~~~~~~~~~~~~~~~batches $\{\langle z_{vk},z_{tk},y_{uvt}\rangle\}_{(v,t)\in \mathcal{X}_{u,:,:}}$ for $u \in \mathcal{U}$, $\{\langle z_{uk},z_{tk},y_{uvt}\rangle\}_{(u,t)\in \mathcal{X}_{:,v,:}}$ for $v \in \mathcal{V}$ and $\{\langle z_{uk},z_{vk},y_{uvt}\rangle\}_{(u,v)\in \mathcal{X}_{:,:,t}}$ for $t \in \mathcal{T}$;\;
\vspace{0.25cm}
\textbf{Procedure} Network\_Parameters\_Update($\mathcal{S}$):\;
~~~~~~~For each $k \in \mathcal{K}~\text{and}~l=1~\text{to}~L$, compute $~~~~~~~~~~~~\scriptsize\nabla_{\boldsymbol{W}^{(\mathcal{S})}_{k,l}} \big[\text{Q}-\text{H}(\boldsymbol{W}^{(\mathcal{S})}_{k,l};\frac{1}{\sigma^{2}})\big] =\;\sum\limits_{s\in\mathcal{S}}\big[\nabla_{\alpha_{sk}}\text{Q}\times\nabla_{\boldsymbol{W}^{(\mathcal{S})}_{k,l}}\alpha_{sk}\scriptsize\big]-\nabla_{\boldsymbol{W}^{(\mathcal{S})}_{k,l}}\text{H}(\boldsymbol{W}^{(\mathcal{S})}_{k,l};\frac{1}{\sigma^{2}})$ with Equations~\ref{eqn:gradient_kl_alpha},~\ref{eqn:gradient_z_alpha}, \ref{eqn:gamma_cdf_pdf} and~\ref{eqn:gamma_cdf_approximate}, and use it to update $\boldsymbol{W}^{(\mathcal{S})}_{k,l}$;\;
~~~~~~~Update $\boldsymbol{b}^{(\mathcal{S})}_{k,l}$, $\boldsymbol{\Phi}^{(\mathcal{S})}_{k,l}$ and $\boldsymbol{\gamma}^{(\mathcal{S})}_{k,l}$ in the same way as $\boldsymbol{W}^{(\mathcal{S})}_{k,l}$ and the updates to $\boldsymbol{\Phi}^{(\mathcal{S})}_{k,l}$ and $\boldsymbol{\gamma}^{(\mathcal{S})}_{k,l}$ are based on Equations~\ref{eqn:gradient_Q_beta},~\ref{eqn:gradient_Q_z},~\ref{eqn:gradient_kl_beta} and~\ref{eqn:gradient_z_beta};\;
\textbf{End Procedure}\; 
\vspace{0.25cm}
\textbf{Procedure} Latent\_Factors\_MCMC\_Sampling($\mathcal{S}$):\;
~~~~~~~For each $k \in \mathcal{K}~\text{and}~s \in \mathcal{S}$, sample $z_{sk} \sim \text{Gamma}(\alpha_{sk},\beta_{sk})$ where $\alpha_{sk}$ and $\beta_{sk}$ are computed based on Equations~\ref{eqn:recursive_fun} and~\ref{eqn:encoder_fun1};\;
\textbf{End Procedure}\;
\vspace{0.25cm}
\textbf{Procedure} Input\_Batches\_Reconstruction($\mathcal{S}_1$, $\mathcal{S}_2$, $\mathcal{S}_3$):\;
~~~~~~~{For each} $k \in \mathcal{K}$, update the input batches for the $k$-th encoders specific to $\mathcal{S}_2$ and $\mathcal{S}_3$ with new samples of $\{z_{sk}\}_{s \in \mathcal{S}_1}$;\;
\textbf{End Procedure}\;
\vspace{0.25cm}
\textbf{Procedure} Mode\_Specific\_Inference($\mathcal{S}_1$, $\mathcal{S}_2$, $\mathcal{S}_3$):\;
~~~~~Network\_Parameters\_Update($\mathcal{S}_1$);\;
~~~~~Latent\_Factors\_MCMC\_Sampling($\mathcal{S}_1$);\;
~~~~~Input\_Batches\_Reconstruction($\mathcal{S}_1,\mathcal{S}_2,\mathcal{S}_3$);\;
\textbf{End Procedure}\;
\vspace{0.25cm}
\textbf{For} \textit{iter}~=~1,~2,~...~\textbf{do}\;
   {
~~~~~Mode\_Specific\_Inference($\mathcal{U}$, $\mathcal{V}$, $\mathcal{T}$); //Inference under mode ``User''\;
~~~~~Mode\_Specific\_Inference($\mathcal{V}$, $\mathcal{U}$, $\mathcal{T}$); //Inference under mode ``Item''\;
~~~~~Mode\_Specific\_Inference($\mathcal{T}$, $\mathcal{U}$, $\mathcal{V}$); //Inference under mode ``Time''\;
}
  \end{algorithm}
\end{figure*}

\section{Inference Scheme of VAE-BPTF}
In this section, we describe how the VAE-BPTF framework infers its parameters. The inference scheme combines the Markov Chain Monte Carlo (MCMC) posterior sampling and the autoencoded variational inference. Algorithm 1 describes the inference scheme in details. In the algorithm, the function $\text{H}( \cdot \text{;}\frac{1}{\sigma^{2}})$ performs regularization on the weights of the encoder networks and the precision $\frac{1}{\sigma^{2}}$ controls its extent. 

As for encoder initialization, Normal distributions $\mathcal{N}(0,\sigma^{2})$ and $\mathcal{N}(0.1,\sigma^{2})$ are used to initialize the output layer weights specific to the posterior shapes and rates\footnote{We found that a small positive mean for the latter Normal distribution could stabilize the algorithm right after the initialization compared to a zero mean.} respectively. As for the hidden layer weights, we use a standard Normal distribution for their initialization. The latent factors under each mode are initialized by the Gamma distribution with the prior shape $\alpha$ and rate $\beta$.

The algorithm conducts inference under each tensor mode, i.e. mode ``User'',  ``Item'' and ``Time''. For each mode-specific inference, the encoder parameters are first updated by the procedure ``Network\_Parameters\_Update'' with Adam optimization~\citep{kingma2014adam}. Then, the latent factors are sampled using the procedure ``Latent\_Factors\_MCMC\_Sampling'' with the updated encoder parameters. Finally, the sampled latent factors under the current mode (e.g. mode ``User'') are used to reconstruct the input batches for the other modes' encoders (i.e. modes ``Item'' and ``Time''). We leverage the ELBO function $\text{Q}$ for testing the convergence of Algorithm 1. If the standard deviation of $\text{Q}$ values over 10 consecutive iterations is sufficiently small, then we deem the algorithm has converged. Otherwise, we terminate the algorithm after 300 iterations. 



\section{Experiments and Results}
The VAE-BPTF framework is evaluated on both synthetic and real-world datasets. The synthetic data evaluation focuses on VAE-BPTF's abilities to recover the right number of latent factors and the posterior parameters. The real-world data evaluation focuses on VAE-BPTF's abilities to reconstruct tensors and generating coherent latent factors. 

\subsection{Synthetic Data Evaluation}
We evaluate the performance of VAE-BPTF using data generated by the framework itself. In this case, we know the number of latent factors and the posterior Gamma shapes and rates for the factors. Therefore, we can compare them with the corresponding estimates from VAE-BPTF.

\begin{figure}[t]
\begin{subfigure}[t]{0.4\textwidth}
\centering
\includegraphics[width=1.5in]{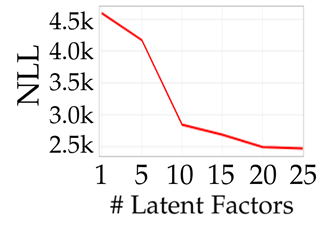}
\caption{}\label{fig:simulation_correlation1}
\end{subfigure}
\hspace{0.3cm}
\begin{subfigure}[t]{0.4\textwidth}
\centering
\includegraphics[width=1.5in]{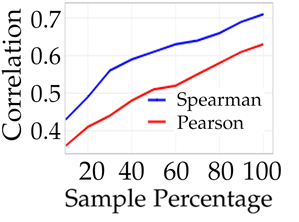}
\caption{}\label{fig:simulation_correlation2}
\end{subfigure}
\caption{~\ref{fig:simulation_correlation1} shows the change of the negative log-likelihood of VAE-BPTF across different numbers of latent factors.~\ref{fig:simulation_correlation2} shows correlations between posterior Gamma parameters and their estimates from VAE-BPTF trained on various percentages of the synthetic data.}
\end{figure}

In particular, we set the number of latent factors to be 10 and the numbers of users, items and time steps to all be 100. As a result, the size of the synthetic tensor is 100$\times$100$\times$100. For each user $u \in \mathcal{U}$, item $v \in \mathcal{V}$ or time step $t \in \mathcal{T}$, we draw their $k$-th latent factors from their respective Gamma distributions as follows: $z_{uk} \sim \text{Gamma}(\alpha_u,\beta_u), z_{vk} \sim \text{Gamma}(\alpha_v,\beta_v), z_{tk} \sim \text{Gamma}(\alpha_t,\beta_t)$. Here, we have $\alpha_u, \alpha_v, \alpha_t, \beta_u, \beta_v, \beta_t \sim \text{Gamma}(\alpha,\beta)$ and set $\alpha=$ 2 and $\beta=$ 0.25. Based on the sampled latent factors, we draw each data entry from Poisson distributions with rates computed using the CP decomposition. The resulting tensor has around 10\% of its data greater than zero. Each MLP encoder of VAE-BPTF is set to have one hidden layer. The parameters $\theta$ and $\eta$ of the reweighting function are set to be 1 and 5 respectively. Figure~\ref{fig:simulation_correlation1} shows that the negative data log-likelihood has a sharp turn on the 10 latent factors. This confirms that VAE-BPTF can recover the optimal number of latent factors (given sufficient data). Figure~\ref{fig:simulation_correlation2} shows that there are positive correlations between the posterior Gamma parameters and their estimates from VAE-BPTF in terms of either Pearson or Spearman coefficient. The positive correlations grow stronger as more data is used to train VAE-BPTF.

\subsection{Real-world Data Evaluation}
We use five real-world datasets to evaluate VAE-BPTF's abilities of tensor reconstruction and generating coherent latent factors. Their domains vary from topic modelling for publication and product reviews, user behaviour modelling for online games, to collaborative filtering on ratings. The following are the descriptions of the datasets:

\begin{itemize}[leftmargin=*]
    \item \textbf{DBLP publication data} (\textbf{DBLP}): This is a four-way count-valued tensor of size 4358 (authors) $\times$ 3308 (venues) $\times$ 4619 (words) $\times$ 52 (years). The data was collected as a random subset of paper abstracts from Semantic Scholar Open Research Corpus\footnote{https://s3-us-west-2.amazonaws.com/ai2-s2-research-public/open-corpus/index.html}. There are 1,444,222 non-zero word counts in the tensor.
    \item \textbf{NIPS publication data} (\textbf{NIPS}): This is a three-way count-valued tensor of size 6427 (authors) $\times$ 4377 (words) $\times$ 14 (years). The tensor was constructed from the abstracts of papers published at the NIPS conference in 2000s\footnote{https://www.kaggle.com/benhamner/nips-papers}. The tensor contains 757,366 non-zero word counts in total.
    \item \textbf{Online game data} (\textbf{Game}): This data was collected and provided by Beijing Shandesitong Technology\footnote{http://www.shandesitong.com/}. A three-way count-valued tensor of size 14064 (users) $\times$ 74 (items) $\times$ 72 (days) was constructed from the data. The tensor records the counts of each virtual item acquired by each user in a Chinese online game on each day over two months. The total number of count values, in this case, is 682,389.
    \item \textbf{Amazon video review data} (\textbf{Video Review}): This is a three-way count-valued tensor of size 5130 (users) $\times$ 1685 (videos) $\times$ 4205 (words). The tensor was built from reviews of Amazon Prime videos\footnote{http://jmcauley.ucsd.edu/data/amazon}. It contains 1,269,654 non-zero word counts.
    \item \textbf{Amazon video rating data} (\textbf{Video Rating}): This is a three-way rating tensor of size 22,088 (users) $\times$ 13,689 (videos) $\times$ 2,465 (days). The data contains 133,087 ratings on Amazon Prime videos\footnote{The prediction targets in this case are the ratings from the Amazon Prime video review dataset.} over a decade.
\end{itemize}

\subsubsection{Reconstruction Error Analysis}
We first evaluate the performance of VAE-BPTF on predicting missing entries in tensors. This is done by comparing VAE-BPTF against five state-of-the-art baselines in terms of both mean absolute error (MAE) and log-likelihood (LL). These baselines are:
\begin{itemize}[leftmargin=*]
    \item \textbf{BPTF}: We applied the version implemented in~\citep{schein2015bayesian}. It uses variational inference to estimate the posterior shapes and rates of latent factors.
    \item \textbf{Online-Gibbs BPTF}: We used the beta-negative binomial version of the BPTF proposed in~\citep{hu2015scalable}. The model is inferred using an online Gibbs sampling scheme.
    \item \textbf{PTF-KL}: We employed the non-Bayesian Poisson tensor factorization model proposed in~\citep{chi2012tensors}. It is optimized over the KL divergence using multiplicative updates.
    \item \textbf{NTF-LS}: We implemented the classical NTF model. The model minimizes the sum of squared (Euclidean) distances using the Adam optimization.
    \item \textbf{MLP-TF}: We extended the MLP-based matrix factorization model \citep{he2017neural} into its tensor version. This new model employs the same objective function and optimization algorithm as the NTF-LS model.
\end{itemize}

We conducted 80-20 training-testing random splitting on each dataset. Then, 5-fold cross-validation was applied to optimize the hyper-parameters of both the baselines and VAE-BPTF. For all the baseline methods, the number of latent factors $|K|$ is selected from the candidate set \{5, 10, 15, 20, 50\}. Specifically for the BPTF model, the shape of the Gamma prior is selected from \{0.1, 1, 10\} and the prior rate is calculated by the default heuristics adopted in its code\footnote{https://github.com/aschein/bptf/blob/master/code/bptf.py}. For the online-Gibbs BPTF model\footnote{https://github.com/ch237/BayesPoissonFactor/blob/master/PTF\_OnlineGibbs.m}, the common hyper-parameter (i.e. denoted by $c$ in the paper) for computing the shapes of the Beta distribution (as a hierarchical prior over the negative binomial model) is selected from \{0.1, 1, 10\}; same as the shape of the Gamma prior (i.e. denoted by $g_r$) in the model. For the PTF-KL model, we adopt the default CP decomposition version from the Tensor Toolbox for Matlab\footnote{https://www.tensortoolbox.org/cp\_apr\_doc.html}. For the NTF-LS model, we implement it with the learning rate selected from \{$10^{-4}$,$10^{-3}$,$10^{-2}$\} and the $\text{L}_2$ regularization term selected from \{$10^{-3}$,$10^{-2}$,...,10\}. Finally, for the MLP-TF model, its number of layers is selected between 1 and 3 and the number of neurons per layer is selected from \{10, 20, 50\}. Its learning rate and $\text{L}_2$ term are selected from the same candidate sets as NTF-LS.

		\begin{table}[h]
		\centering\setlength\tabcolsep{4pt}
		\begin{tabular}{|c|c|c|c|c|c|c|}				\hline
			Data&$|K|$ &$L$&$d_{l}$&h, q& $\theta$, $\eta$&$\sigma^{2}$\\	
			\hline
			\multicolumn{7}{|c|}{\textbf{Full Data}}\\
			\hline
			DBLP&10 &1&20&\scriptsize{Softplus, Softplus}& 5, 10&1\\
			NIPS&10 &1&20&\scriptsize{Softplus, Softplus}& 5, 10&5\\
			Game&5 &1&10&\scriptsize{Softplus, Softplus}& 5, 10&1\\
			Review&5 &1&10&\scriptsize{Softplus, Softplus}& 10, 50&1\\
			Rating&15 &2&20&\scriptsize{Softplus, ReLU}& 3, 10&5\\
			\hline
			\multicolumn{7}{|c|}{\textbf{Sub-sampled Data}}\\
			\hline			DBLP&3 to 5 &0 to 1&10&\scriptsize{Softplus, Softplus}& 5, 20&0.1/1\\
			NIPS&3 to 5 &0 to 1&10&\scriptsize{Softplus, Softplus}& 5, 20&0.1/1\\
			Game&3 &0 to 1&10&\scriptsize{Softplus, Softplus}& 5, 20&0.1\\
			Review&3 to 5 &0 to 1&10&\scriptsize{Softplus, Softplus}& 10, 50&0.1/1\\
			Rating&3 to 10 &0 to 1&10&\scriptsize{Softplus, ReLU}& 5, 10&1\\
			\hline
		\end{tabular}
		\caption{Hyper-parameter optimization results for VAE-BPTF over the five datasets and their respective sub-sampled datasets. For the sub-sampled datasets, the hyper-parameters can take multiple values (e.g. 3 to 5 for $|K|$) corresponding to different sample percentages.}
		\label{table:hyper_param_optim}
	\end{table}

The hyper-parameters of VAE-BPTF include the number of latent factors $|K|$, the number of encoder hidden layers $L$, the number of neurons per layer $d_{l}$, the selection of activation functions h and q, the reweighting parameters $\theta$ and $\eta$, and the variance $\sigma^2$. Table~\ref{table:hyper_param_optim} shows the hyper-parameter optimization results for VAE-BPTF over both the five datasets and their sub-sampled datasets. Note that the optimized values for the prior shape $\alpha$ and the prior rate $\beta$ are not included in the table. We found that their values, which are both 1, are overall insensitive to the datasets and the selection of the other hyper-parameters in terms of MAE and LL. As for the value of $\bar{y}$, we used the most frequent target value of each training dataset, that is 5 for the Amazon rating data and 1 for all the other datasets. From Table~\ref{table:hyper_param_optim}, we can see that VAE-BPTF needs more latent factors and hidden layers to fit the Amazon rating data. Moreover, it is the only dataset for which VAE-BPTF uses the ReLU function in the hidden layers. It is also observed that the reweighting parameter values for the Amazon review data are much higher than those for the other datasets. This suggests sharper increases in the reweight terms from the most frequent count value to less frequent values. As a result, data instances with count values close to $\bar{y}$ will still be considered important. Finally, it can be observed that the values for the hyper-parameters are smaller when VAE-BPTF fits the sub-sampled datasets. In addition, greater regularization (i.e. greater values for $\frac{1}{\sigma^2}$) is also exerted by VAE-BPTF on the model parameters.
	\begin{table*}[h]
		\centering
		\begin{tabular}{|c|c|c|c|c|c|}
			\hline
			\multirow{2}{*}{Models}&&&& \multicolumn{2}{|c|}{Video}\\
			&DBLP&NIPS&Game&Review&Rating\\
			\hline
			VAE-BPTF&\textbf{0.656}&\textbf{0.532}&\textbf{1.241}&\textbf{0.226}&\textbf{0.756}\\
			\hline
            BPTF&0.798&0.914&1.558&0.454&1.447\\
			\hline
			Online-Gibbs&\multirow{2}{*}{1.412}&\multirow{2}{*}{1.183}&\multirow{2}{*}{1.575}&\multirow{2}{*}{1.207}&\multirow{2}{*}{3.472}\\
			\hhline{~~~~~}
			BPTF&&&&&\\
			\hline
			PTF-KL&1.521&1.223&1.625&1.211&4.023\\
			\hline
			NTF-LS&1.144&1.343&1.896&1.217&3.443\\
			\hline
			MLP-TF&0.728&0.616&1.387&0.362&0.770\\
			\hline

		\end{tabular}
		\caption{The mean absolute error of each model on the different datasets (the best performance shown in bold figures)}
		\label{table:mae_results}
	\end{table*}

	\begin{table*}[h]
		\centering
		\begin{tabular}{|c|c|c|c|c|c|}
			\hline
			\multirow{2}{*}{Models}&&&& \multicolumn{2}{|c|}{Video}\\
			&DBLP&NIPS&Game&Review&Rating\\
			\hline
			VAE-BPTF&\textbf{-3.05}$\boldsymbol{\times10^{5}}$&\textbf{-1.31}$\boldsymbol{\times10^{5}}$&\textbf{-3.84}$\boldsymbol{\times10^{4}}$&\textbf{-2.29}$\boldsymbol{\times10^{5}}$&\textbf{6.28}$\boldsymbol{\times10^{4}}$\\
			\hline
			BPTF&-8.47$\times10^{5}$&-4.25$\times10^{5}$&-1.43$\times10^{5}$&-6.53$\times10^{5}$&2.26$\times10^{4}$\\
			\hline
			Online-Gibbs&\multirow{2}{*}{-1.38$\times10^{6}$}&\multirow{2}{*}{-5.87$\times10^{5}$}&\multirow{2}{*}{-1.62$\times10^{5}$}&\multirow{2}{*}{-2.17$\times10^{6}$}&\multirow{2}{*}{-3.52$\times10^{5}$}\\
			\hhline{~~~~~}
			BPTF&&&&&\\
			\hline
			PTF-KL&-3.29$\times10^{6}$&-7.26$\times10^{5}$&-2.33$\times10^{5}$&-2.64$\times10^{6}$&-6.11$\times10^{5}$\\
			\hline
		\end{tabular}
		\caption{The log-likelihood of each model on the different datasets (the best performance shown in bold figures)}
		\label{table:ll_results}
	\end{table*}
	
				\begin{table}[h]
		\centering
		\begin{tabular}{|c|c|c|c|c|}
			\hline
			\multirow{3}{*}{Data}&\multicolumn{2}{|c|}{}&\multicolumn{2}{|c|}{VAE-BPTF without}\\
			&\multicolumn{2}{|c|}{VAE-BPTF}&\multicolumn{2}{|c|}{Reweights}\\
			\hhline{~----}
			& MAE&LL&MAE&LL\\
			\hline
			DBLP&0.656&-3.05$\times10^{5}$&0.697&-4.33$\times10^{5}$\\
			\hline
			NIPS&0.532&-1.31$\times10^{5}$&0.578&-2.10$\times10^{5}$\\
			\hline
			Game&1.241&-3.84$\times10^{4}$&1.274&-5.63$\times10^{4}$\\
			\hline
			Review&0.226&-2.29$\times10^{5}$&0.260&-3.52$\times10^{5}$\\
			\hline
			Rating&0.756&6.28$\times10^{4}$&0.772&5.48$\times10^{4}$\\
			\hline
		\end{tabular}
		\caption{Ablation study results for VAE-BPTF on the different datasets}
		\label{table:ablation_results}
	\end{table}

Tables~\ref{table:mae_results} and~\ref{table:ll_results} respectively show the mean absolute error and the log-likelihood of different models on each dataset. It can be observed that VAE-BPTF has outperformed the baseline models across all the datasets in terms of both metrics. Its superiority over the other Poisson factorization models is obvious, especially with an increase of one order of magnitude in the log-likelihood. This demonstrates that VAE-BPTF is much better at inferring the posterior distributions of the latent factors. As a result, there is much less uncertainty about its predictions being closer to the ground-truth values as measured by the log-likelihood. 

In addition, the online-Gibbs BPTF model, which focuses on modelling over-dispersed count data with a beta-negative binomial construction, has failed to outperform even the BPTF model on the three text analysis datasets which are the DBLP, NIPS and Video Review. We conjecture that this is because the extent of word burstiness \citep{Buntinetopic2014} that causes over-dispersed word counts is not significant enough in these datasets. This is also evidenced by the fact that its performance is much worse on the DBLP dataset than on the NIPS dataset as the former has one more tensor mode which diffuses the word burstiness even more. On the other hand, VAE-BPTF appears to be least affected by the over-dispersion, if there is any. This might be partially attributed to its reweighting scheme that essentially penalizes high variance in word counts (as those down-weighted low-count words can be viewed as being discarded from the bag of words).  

The baseline closest to VAE-BPTF in performance is the MLP-TF model. Its performance was optimized with 20 embedding dimensions\footnote{The embedding was done based on the entity IDs.} (for each tensor mode), 2 hidden layers with 50 ReLU neurons each and a negative sampling ratio of 3:1\footnote{Three zero values per one non-zero values.}. Nonetheless, the superiority of VAE-BPTF over MLP-TF is statistically significant according to a one-tailed paired t-test where the p-value equals 0.025. The results also suggest that the efficacy of incorporating neural components into traditional factorization models appears to outweigh the efficacy of intricate probabilistic modelling. In addition, we did not compute the log-likelihoods for both MLP-TF and NTF-LS. This is because their squared loss function does not include a standard deviation term during its optimization.

\begin{figure*}[t]
\centering
\begin{subfigure}[t]{0.5\textwidth}
\includegraphics[width=1.75in]{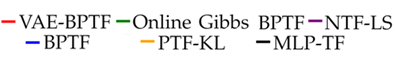}
\end{subfigure}\\
\begin{subfigure}[t]{0.5\textwidth}
\centering
\includegraphics[width=3in]{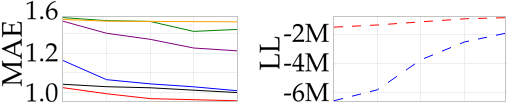}
\caption{DBLP}\label{fig:performance_dblp_subsample}
\end{subfigure}\\
\begin{subfigure}[t]{0.5\textwidth}
\centering
\includegraphics[width=3in]{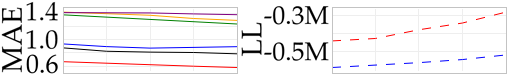}
\caption{NIPS}\label{fig:performance_nips_subsample}
\end{subfigure}\\
\begin{subfigure}[t]{0.5\textwidth}
\centering
\includegraphics[width=3in]{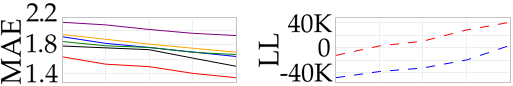}
\caption{Game}\label{fig:performance_dblp_subsample}
\end{subfigure}\\
\begin{subfigure}[t]{0.5\textwidth}
\centering
\includegraphics[width=3in]{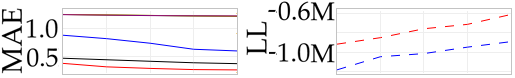}
\caption{Amazon Review}\label{fig:performance_amazon_subsample1}
\end{subfigure}\\
\begin{subfigure}[t]{0.5\textwidth}
\centering
\includegraphics[width=3in]{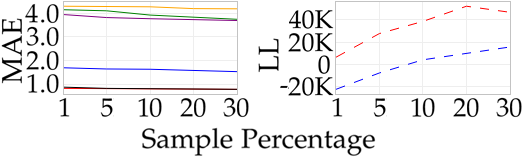}
\caption{Amazon Ratings}\label{fig:performance_amazon1_subsample1}
\end{subfigure}
\caption{The mean absolute error and log-likelihood of each model on sub-sampled datasets; The sample percentages are 1\%, 5\%, 10\%, 20\% and 30\%.}\label{fig:result_subsampled_data}
\end{figure*}

Table~\ref{table:ablation_results} shows the result of the ablation study on VAE-BPTF in which its reweighting scheme was taken away. Its performance was degraded accordingly across all the datasets. This result suggests the importance of properly handling the imbalanced data values. Furthermore, without the reweighting scheme, VAE-BPTF still notably outperforms the other baselines. This indicates that despite using the CP decomposition (as the other baseline models) as the decoder, VAE-BPTF's variational encoders are better at predicting the posterior parameters for the latent factors. We also observed that without the reweighting scheme, VAE-BPTF spent more iterations reaching its lowest MAE and highest LL on the validation datasets.

As a further evaluation, we would like to see how robust our framework is towards the data sparsity issue, i.e., small numbers of count values per entity under each mode. This problem is prevalent in the cold-start scenario~\citep{rashid2008learning}. For the evaluation, we need to run each model on small to medium data subsets and observe the corresponding trends in their performance. To obtain these data subsets, we sub-sampled the original datasets by 1\%, 5\%, 10\%, 20\% and 30\% (irrespective of entities under different tensor modes). When the subsampling percentage is 1\%, the number of observed entries left in the tensor of each dataset is no less than thousands, which ensures that all the models have enough data to be trained and validated properly. We ran VAE-BPTF on the resulting data subsets with the 5-fold cross-validation for optimization of the hyper-parameters. The results are specified in Table~\ref{table:hyper_param_optim}. For the baseline models, their hyper-parameters were optimized under the same validation strategy from the candidate values specified at the beginning of this section.

Figure~\ref{fig:result_subsampled_data} shows that VAE-BPTF achieved overall lower MAE and higher LL\footnote{We show only the LLs of VAE-BPTF and BPTF as the other Poisson-based models are significantly inferior to them in this aspect.} than the other models. VAE-BPTF also exhibits smooth trends in both metrics between 1\% and 10\% of each dataset. This suggests that the complexity of VAE-BPTF's network structure has been properly regularized by the validation strategy. As a result, its network parameters can be reliably learned from the sparse data.
\begin{table}[h]
		\centering
		\begin{tabular}{|c|c|c|c|c|c|}
			\hline
					NPMI&DBLP &NIPS& Video\\
			&&&Review\\
			\hline
	VAE-BPTF&\textbf{-0.161}&\textbf{-0.145}&\textbf{-0.202}\\
			\hline
	BPTF&-0.226&-0.174&-0.241\\
			\hline
	Online-Gibbs BPTF&-0.245&-0.211&-0.249\\
			\hline
	PTF-KL&-0.236&-0.198&-0.261\\
			\hline
	NTF-LS&-0.247&-0.224&-0.267\\
			\hline	MLP-TF&-0.193&-0.206&-0.228\\
			\hline
		\end{tabular}
		\caption{The average NPMI scores of each model on the words from the corpus of each dataset (the best performance shown in bold figures)}
		\label{table:results}\label{table:npmi_results}
	\end{table}

\subsubsection{Latent Factor Coherence Analysis}
We further evaluate the semantic coherence of latent factors specific to words. We first computed the adjacency matrix of word latent factors in terms of the Euclidean distance. According to this matrix, we found the top 10 words closest to each word (including the word itself). Then, for each word, we used the Normalized Point-wise Mutual Information (NPMI)~\citep{aletras2013evaluating} to calculate a coherence score for its top 10 words\footnote{The NPMI scoring uses a large Wikipedia dump hosted by Palmetto: http://palmetto.aksw.org.}. A higher score indicates greater coherence among the top 10 words. We removed the scores of rare words (i.e. words that occurred in less than 0.1\% of the documents in the corpus of each dataset). Finally, we averaged the scores across the remaining words and the results are summarized in Table~\ref{table:npmi_results} \footnote{The Game data and the Amazon rating data are not text data, and thus NPMI is not applicable.}. It shows that VAE-BPTF achieved greater average semantic coherence on words compared to the other models. Moreover, the coherence degrees of the models are overall consistent with their tensor reconstruction performance. This indicates that the tensors have been generated coherently and VAE-BPTF explains this coherent generation more effectively.

\begin{figure*}[t]
\begin{minipage}[c][5.75cm][t]{.5\textwidth}
  \vspace*{\fill}
  \centering
  \includegraphics[width=2.25in]{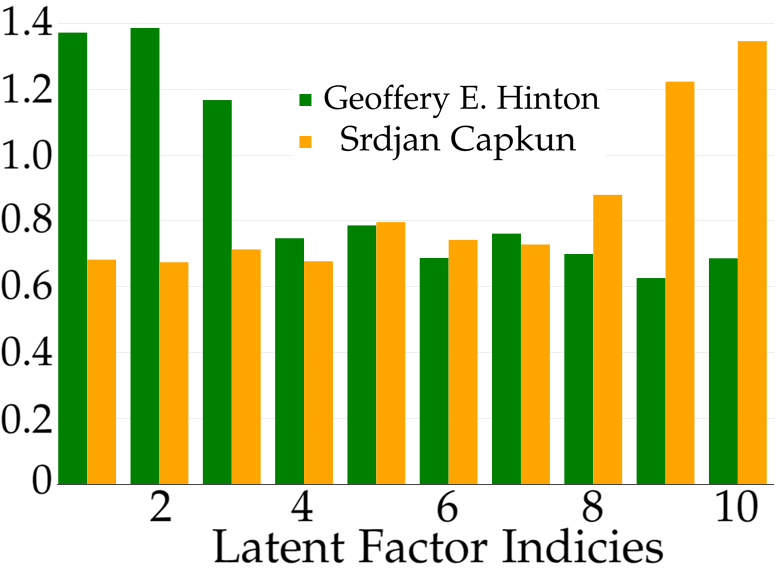}
  \subcaption{Latent factors of two authors on DBLP who respectively specialize in neural networks and information security}
  \label{fig:latent_factors_plots_DBLP}
\end{minipage}%
\begin{minipage}[c][6.25cm][t]{.5\textwidth}
\vspace*{\fill}
  \centering
  \includegraphics[width=2.25in]{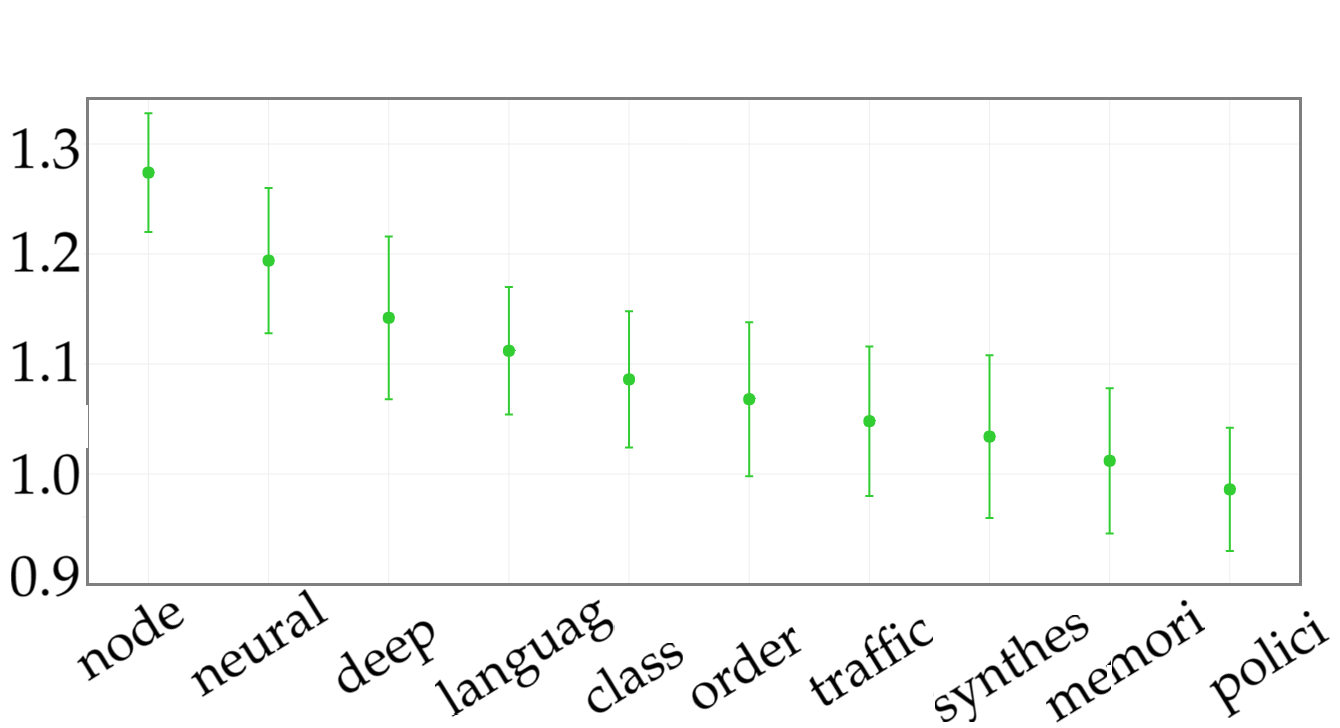}
  \par\vfill
  \includegraphics[width=2.75in]{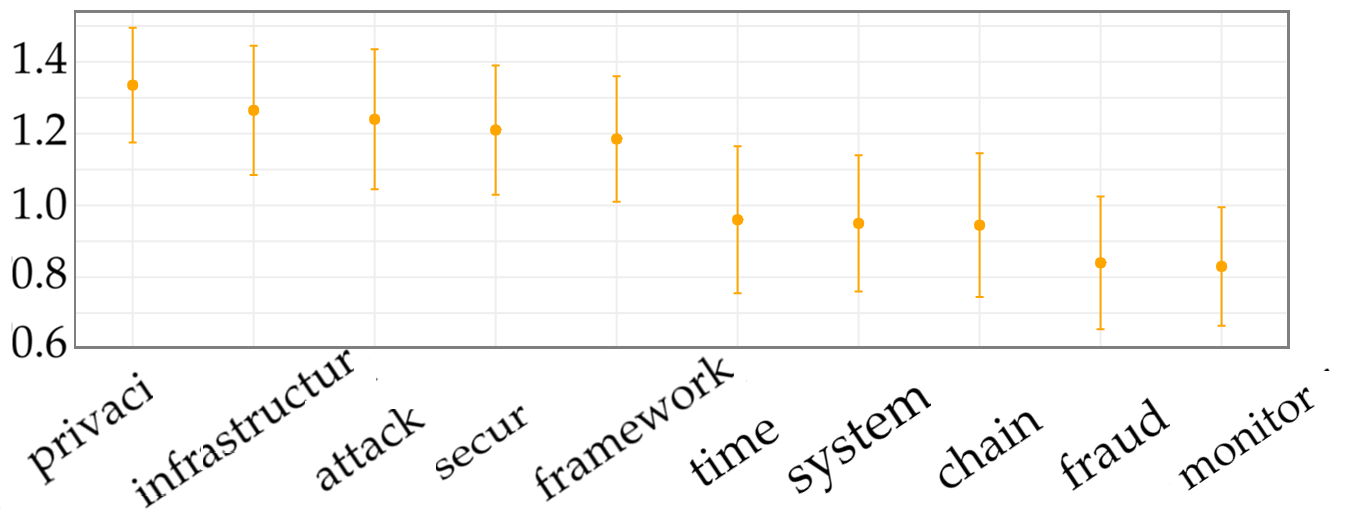}
  \subcaption{The authors' top 10 words in use}
  \label{fig:latent_factors_plots_DBLP_top10}
\end{minipage}
\caption{Qualitative analysis of the latent factors generated by VAE-BPTF on the DBLP dataset.}\label{fig:latent_factors_plots}
\end{figure*}
\subsubsection{Qualitative Analysis}
We further inspect the latent factors generated by VAE-BPTF on the DBLP and Amazon review datasets. Figure~\ref{fig:latent_factors_plots_DBLP} shows the latent factors of two authors on DBLP who specialize in different research areas. Different spikes in their latent factors indicate that they prefer to use different (groups of) words that have the corresponding latent factor patterns. This is reconfirmed by Figure~\ref{fig:latent_factors_plots_DBLP_top10} which shows the authors' respective top 10 words in use. To acquire these words, we first computed the dot products between the samples of each author's and each word's posterior mean latent factors. More specifically, the samples were collected as the means of the posterior Gammas from the VAE-BPTF encoders over 50 iterations after convergence. Then, the dot product results were averaged over the 50 samples and the averages were sorted in descending order per author. Finally, the top 10 words were selected per author according to the sorted values. Figure~\ref{fig:latent_factors_plots_DBLP_top10} also shows the standard deviations of the dot product results across the samples. It can be observed that the top 10 words inferred by VAE-BPTF are directly relevant to the research areas of the two authors: neural networks and information security.

\begin{figure*}
\begin{minipage}[c][6.75cm][t]{.5\textwidth}
  \vspace*{\fill}
  \centering
  \includegraphics[width=2.2in]{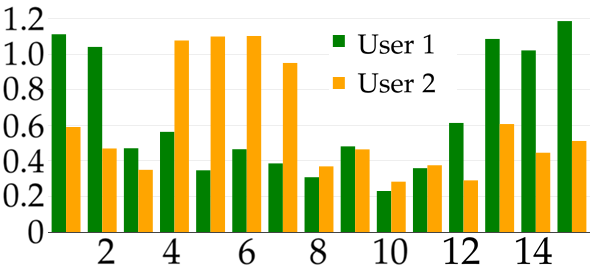}
  \subcaption{Latent factors of two Amazon users who respectively prefer crime dramas and animated comedies}  \label{fig:latent_factors_plots_Amazon}
    \includegraphics[width=2.3in]{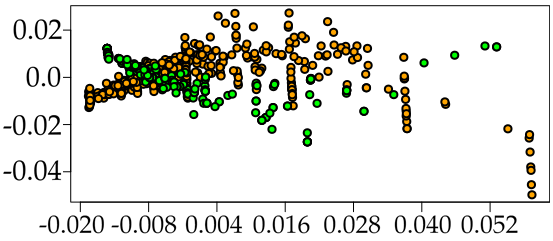}
  \subcaption{Embeddings of latent factors of two video genres: crime (green) and comedy (orange)}  \label{fig:latent_factors_plots_scaling}
\end{minipage}%
\begin{minipage}[c][5.5cm][t]{.5\textwidth}
  \includegraphics[width=2.8in]{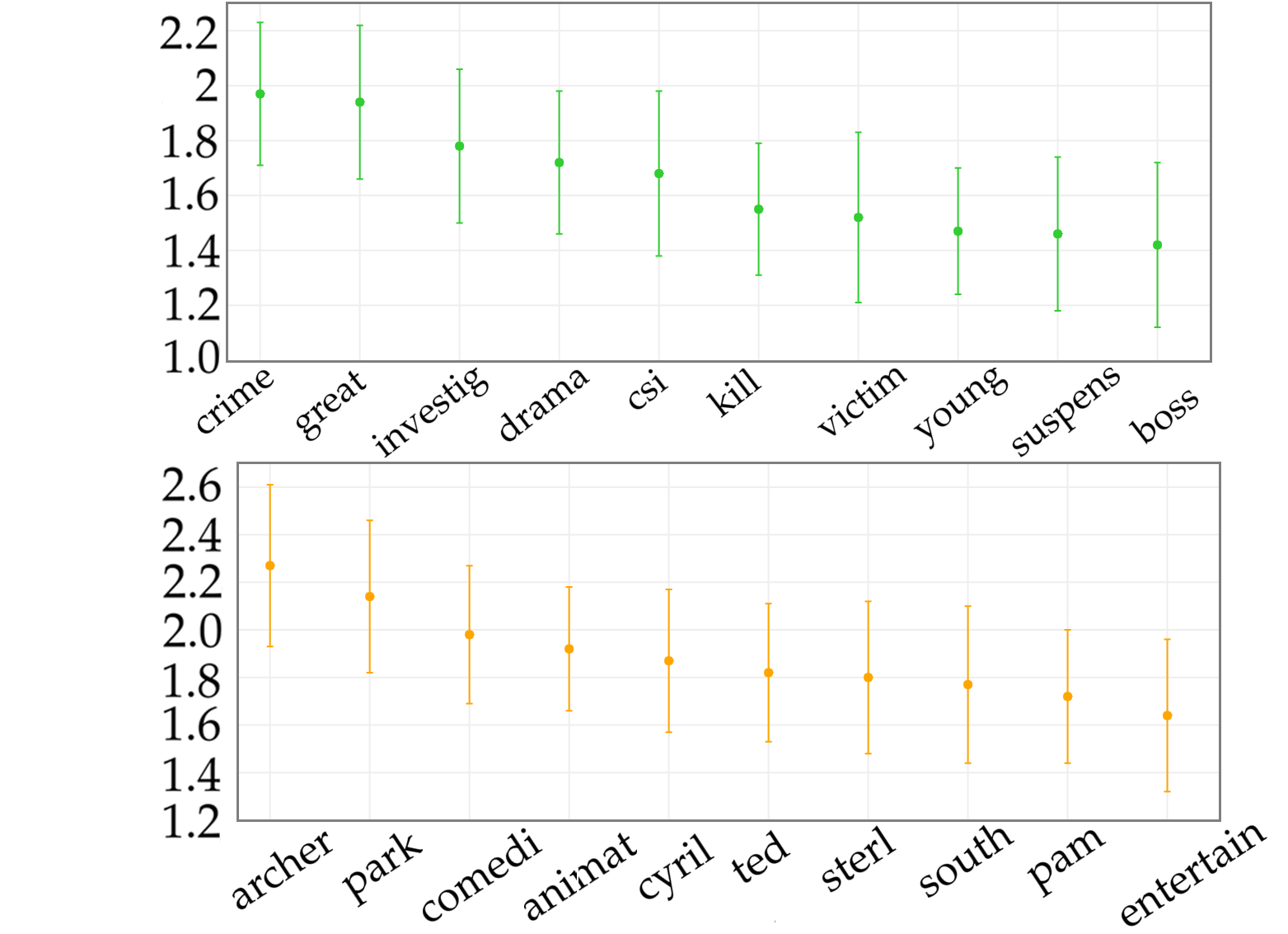}
  \subcaption{The users' top 10 words in use} \label{fig:latent_factors_plots_Amazon_top10}
\end{minipage}
\caption{Qualitative analysis of the latent factors generated by VAE-BPTF on Amazon video data}
\end{figure*}

Likewise, Figure~\ref{fig:latent_factors_plots_Amazon} shows the difference in the latent factors of two Amazon users who prefer either crime dramas or (animated) comedies. Figure~\ref{fig:latent_factors_plots_Amazon_top10} further displays the top 10 words in use in their respective reviews. In addition, Figure~\ref{fig:latent_factors_plots_scaling} shows two-dimensional embeddings of the latent factors of crime and comedy videos. To obtain the embeddings, we applied multi-dimensional scaling\footnote{We used the cmdscale function in R that implements the classical multi-dimensional scaling.}~\citep{cox2000multidimensional} to the latent factors of videos with single tags that are either crime or comedy. From the figure, a notable difference can be observed in the scatters of the embeddings of two video genres. This is coherent with the human perception that there should be some difference in the two genres.


\subsubsection{Computational Complexity Analysis}
The computational complexity of our framework is approximately $\mathcal{O}\big(N\times M\times K\times \sum^{L+1}_{l=0}d_{l}d_{l+1}\big)$, where $N$ is the number of non-zero entries in the tensor, $M$ is the number of tensor modes and $d_{l}d_{l+1}$ is the number of weights at the $l$-th hidden layer. In theory, our framework is computationally more expensive than the BPTF model whose complexity is $\mathcal{O}\big(N\times M\times K\big)$. In practice, we found that the running time of our framework was generally an order of magnitude slower than the BPTF model, given that $d_{l}d_{l+1}$, $K$ and $L$ can be directly obtained from Table 1 across the experiment datasets. 

We have also tried a lighter version of VAE-BPTF whose $K$ encoders under the same mode share the same set of weights. This variant has a computational complexity of $\mathcal{O}(N\times M\times \sum^{L+1}_{l=0}d_{l}d_{l+1})$. However, this variant has performed notably worse than the original framework across the experiment datasets. Nevertheless, the computational complexity remains to be a limitation of our framework that needs to be addressed in the future work.

\section{Conclusion and Future Work}
In this paper, we proposed the VAE-BPTF framework which integrates non-negative tensor factorization with variational auto-encoders. The encoder networks compute the posterior Gamma parameters for each latent factor specific to entities under each tensor mode. More specifically, a parameter is computed by summing the softplus activation of the encoder outputs. Each output is computed via an MLP network. An input to this MLP comprises a data value generated by the target entity and the corresponding latent factors of the other entities. Furthermore, to deal with the imbalance problem in count data, VAE-BPTF downweighs the softplus activation of those corresponding to common data values. 

According to the synthetic data evaluation, VAE-BPTF could find the right number of latent factors and accurately estimate the posterior parameters. Moreover, VAE-BPTF outperformed state-of-the-art tensor factorization models on five real-world datasets in terms of reconstruction errors and latent factor coherence. We conducted qualitative analysis on the inferred latent factors of different entities and found that they tend to agree with the human.

For future work, we would like to experiment with reweighting schemes that consider weighted errors. They assign more weights to softplus values corresponding to data values that were predicted less accurately in previous rounds. Furthermore, we can investigate whether ensemble learning can be incorporated into VAE-BPTF. For example, boosting techniques can be applied to sequentially build weak encoder networks. We expect the base network to account for the imbalance of data values and the following networks to improve the fit on the residuals. Another research direction is to develop variants of VAE-BPTF that aim to reduce the computational complexity and meanwhile, maintain comparable prediction performance and degrees of coherence in generated latent factors. 

\bibliographystyle{spbasic}

\bibliography{bibliography}

\end{document}